\documentclass{article}

\usepackage[preprint]{neurips_2026}
\usepackage{xspace}
\newcommand{\ie}{\textit{i.e.,}\xspace}

\usepackage[utf8]{inputenc}
\usepackage[T1]{fontenc}
\usepackage{url}
\usepackage{booktabs}
\usepackage{amsfonts}
\usepackage{nicefrac}
\usepackage{microtype}
\usepackage{xcolor}
\usepackage{graphicx}
\usepackage{float}
\usepackage{wrapfig}
\usepackage{amsmath,amssymb}
\usepackage{xcolor}
\definecolor{lightpurple}{RGB}{180,150,255}
\usepackage{kotex}
\usepackage{multirow}
\usepackage{titlesec}
\usepackage{caption} 
\titlespacing*{\paragraph}
  {0pt}      
  {-1mm}     
  {0.5em}    

\usepackage[
    colorlinks=true,
    citecolor=lightpurple,
    linkcolor=lightpurple,
    urlcolor=lightpurple
]{hyperref}

\RequirePackage[normalem]{ulem} 
\RequirePackage{xcolor} 
\providecommand{\DIFaddbegin}{} 
\makeatletter 
\let\sout@orig\sout 
\renewcommand{\sout}[1]{\ifmmode\text{\sout@orig{\ensuremath{#1}}}\else\sout@orig{#1}\fi} 
\makeatother 

\title{SkillRet: A Large-Scale Benchmark \\ for Skill Retrieval in LLM Agents}
\author{
  Ryangkyung Kang\textsuperscript{1}\thanks{Equal contribution.}
  \quad
  Hongcheol Cho\textsuperscript{1}\footnotemark[1]
  \quad
  Youngeun Kim\textsuperscript{2}\thanks{Corresponding author: \texttt{youngeunkim@korea.ac.kr}}
  \\
  \\
  \textsuperscript{1}ThakiCloud
  \quad
  \textsuperscript{2}Korea University
}

\begin{document}

\maketitle

\begin{abstract}
As LLM agents are increasingly deployed with large libraries of reusable skills, selecting the right skill for a user request has become a critical systems challenge. In small libraries, users may invoke skills explicitly by name, but this assumption breaks down as skill ecosystems grow under tight context and latency budgets. Despite its practical importance, skill retrieval remains underexplored, with limited benchmarks and little understanding of retrieval behavior on realistic skill libraries. To address this gap, we introduce \textsc{SkillRet}, a large-scale benchmark for skill retrieval in LLM agents. \textsc{SkillRet} contains 16,129 public agent skills, organized with structured semantic tags and a two-level taxonomy spanning 6 major categories and 18 sub-categories. It provides 63,259 training samples and  4,392 evaluation queries with disjoint skill pools,enabling both benchmarking and retrieval-oriented training.
Across a diverse set of retrievers, we find that skill retrieval remains far from solved: off-the-shelf models struggle on realistic large-scale skill libraries, and prior skill-retrieval models still leave substantial headroom. Task-specific fine-tuning on \textsc{SkillRet} improves NDCG@10 by +12.9 points over the strongest prior retriever and by +16.2 points over the strongest off-the-shelf retriever. Our analysis further suggests that these gains arise because fine-tuned models better focus on the small skill-relevant signals within long and noisy queries. These results establish \textsc{SkillRet} as a strong benchmark and foundation for future research on retrieval in large-scale agent systems.
We publicly release the \href{https://huggingface.co/datasets/ThakiCloud/SKILLRET}{benchmark}, \href{https://github.com/ThakiCloud/SKILLRET}{code}, and model checkpoints (\href{https://huggingface.co/ThakiCloud/SKILLRET-Embedding-0.6B}{0.6B}, \href{https://huggingface.co/ThakiCloud/SKILLRET-Embedding-8B}{8B}). \end{abstract}


\section{Introduction}
\label{sec:introduction}
As LLM agents become more capable, they increasingly rely on reusable skills (\ie long-form procedural modules such as prompts, scripts, workflows, and execution policies) to solve complex tasks \cite{xu2026agent,jiang2026sok,zhou2026memento,wang2023voyager}. In small-scale settings, users can often invoke such skills explicitly by name. However, this assumption becomes brittle as agent ecosystems grow. When a system maintains a large default pool of reusable skills, it is no longer practical to expose the entire library in context or expect users to know which skill should be activated for a given request. Instead, future agent systems will increasingly require an explicit retrieval layer that selects a small, relevant subset of skills for the current task, both to reduce context cost and to enable robust automated skill use at scale \cite{li2025skillflow}. This shift is already visible in recent agent systems such as MetaClaw~\cite{xia2026metaclaw}, XSkill~\cite{jiang2026xskill}, and WebXSkill~\cite{wang2026webxskill}, which rely on inference-time retrieval of task-relevant skills or knowledge to guide downstream execution.

This trend makes skill retrieval and selection a central systems problem. The key challenge is whether agents can identify the right skills from a large library under realistic inference constraints. However, despite the growing need for reliable skill selection, its evaluation remains underdeveloped. As shown in Table~\ref{tab:intro_comparison}, prior skill benchmarks \cite{li2026skillsbench,han2026sweskillsbench,li2026agentskillos} mainly focus on end-to-end execution rather than retrieval itself, while existing retrieval benchmarks either target tools or provide only limited evaluation scale. ToolRet studies tool retrieval and shows that even strong IR models struggle in that setting \cite{shi2025toolret}. SkillRouter is the closest prior work on skill retrieval, but provides only 75 evaluation queries and does not publicly release its training data \cite{zheng2026skillrouter}. These limitations point to the need for a larger, publicly available benchmark with substantial training and evaluation splits that isolates skill retrieval as a standalone problem.

To address this gap, we introduce \textsc{SkillRet}, a large-scale benchmark for skill retrieval in LLM agents. \textsc{SkillRet} is built from 16{,}129 public agent skills, curated from a raw crawl of 22{,}795 listings through a filtering pipeline. It provides 63{,}259 public training samples and 4,392 evaluation samples, enabling both controlled benchmarking and retrieval-oriented model development. We further annotate the corpus with semantic tags and a two-level taxonomy spanning 6 major categories and 18 sub-categories, supporting fine-grained analysis across domains and difficulty factors. Altogether, \textsc{SkillRet} captures a realistic retrieval environment characterized by long-context skill documents and imbalanced skill distributions.

We benchmark a broad range of retrieval and reranking models on \textsc{SkillRet}. Our experiments reveal several key findings. First, skill retrieval remains challenging: even the strongest off-the-shelf retriever achieves limited performance, indicating that existing models are not well suited for retrieving relevant skills from queries. Second, task-specific fine-tuning on our training data yields substantial gains, allowing smaller fine-tuned models to match or even surpass much larger off-the-shelf models. Third, reranking is most effective when the first-stage retriever has remaining headroom, but its marginal benefit diminishes once the base retriever becomes strong. Also, our analysis shows that fine-tuned models improve retrieval by better focusing on the small skill-relevant sentences embedded within long, noisy, and compositional queries. Finally, the benchmark signal transfers beyond synthetic labels. On Terminal-Bench \cite{merrill2026terminal}, retrieved skills preserve agent success while reducing cost by 10\%. These results establish skill retrieval as a distinct retrieval problem and position \textsc{SkillRet} as a strong foundation for future research in large-scale agent systems.

\begin{table}[t]
  \centering
  \caption{%
  Comparison of \textsc{SkillRet} with related benchmarks and work.
  Unlike prior skill benchmarks that mainly evaluate end-to-end performance, \textsc{SkillRet} isolates skill retrieval as a standalone problem and provides large-scale train/evaluation splits for retrieval-model development.
  Compared with the closest skill-retrieval work, SkillRouter, \textsc{SkillRet} offers a substantially larger evaluation set and a larger public training set.
  $^\dagger$SkillRouter reports 37,979 training data, but the training data are not publicly released.
}
  \label{tab:intro_comparison}
  \small
  \setlength{\tabcolsep}{5pt}
  \resizebox{\columnwidth}{!}{%
  \begin{tabular}{l l c c c c}
  \toprule
  \textbf{Benchmark} & \textbf{Task} & \textbf{Target}
    & \textbf{\# Eval Samples}
    & \textbf{Train}
    & \textbf{\# Train Samples} \\
  \midrule
  ToolRet~\cite{shi2025toolret}
    & Retrieval & Tool & 7,615
    & \checkmark & $>$200K \\
  \midrule
  SkillsBench~\cite{li2026skillsbench}
    & End-to-End Performance & Skill & 86
    & $\times$ & -- \\
  SWE-Skills-Bench~\cite{han2026sweskillsbench}
    & End-to-End Performance & Skill & 565
    & $\times$ & -- \\
  AgentSkillOS~\cite{li2026agentskillos}
    & End-to-End Performance & Skill & 30
    & $\times$ & -- \\
  \midrule
  SkillRouter~\cite{zheng2026skillrouter}
    & Retrieval & Skill & 75
    & \checkmark & 37,979$^\dagger$ \\
  \textbf{\textsc{SkillRet} (ours)}
    & \textbf{Retrieval} & \textbf{Skill} & \textbf{4,392}
    & \checkmark & \textbf{63,259} \\
  \bottomrule
  \end{tabular}%
  }
\end{table}

\section{Related Work}
\label{sec:relatedwork}

\subsection{Agent Skills}
\label{sec:rw-skill-agents}

Recent work {increasingly} treats skills as reusable modules and increasingly makes them a core component of agent design.
MetaClaw proposes a continual meta-learning framework that jointly evolves a base LLM policy and a reusable skill library, using failure trajectories to synthesize new skills and improve agents without downtime \cite{xia2026metaclaw}.
XSkill studies continual learning in multimodal agents through two forms of reusable knowledge retrieved and adapted to the current visual context at inference time \cite{jiang2026xskill}.
WebXSkill focuses on autonomous web agents and introduces executable skills that combine parameterized action programs with step-level natural language guidance, organized in a URL-based graph for context-aware retrieval \cite{wang2026webxskill}.
These systems show that reusable skills are becoming a practical design pattern and that inference-time access to skill libraries is increasingly important.
Another research direction studies broader skill ecosystems and downstream usefulness.
AgentSkillOS studies ecosystem-scale organization, selection, and orchestration through capability trees and DAG-based multi-skill pipelines, evaluating 30 artifact-rich tasks across five categories \cite{li2026agentskillos}.
SkillsBench measures whether skills improve performance across 86 tasks in 11 domains, showing gains from curated skills but no average benefit from self-generated skills \cite{li2026skillsbench}.
SWE-Skills-Bench similarly evaluates public SWE skills on requirement-driven software engineering tasks and finds that most skills provide little or no pass-rate improvement \cite{han2026sweskillsbench}.
These works complement ours by showing that skill ecosystems are already emerging and that downstream skill usefulness is highly variable. However, these benchmarks do not isolate skill retrieval quality as a standalone problem. In end-to-end skill-use settings, failures can arise from the intrinsic usefulness of the selected skill, orchestration errors, execution failures, or contextual mismatch, making it difficult to attribute performance specifically to retrieval.

\subsection{Skill Retrieval Benchmarks and Skill Routing}
\label{sec:rw-retrieval}

A smaller but growing line of work studies retrieval more directly.
In the tool setting, ToolRet introduces a benchmark with 7.6K retrieval tasks and 43K tools, showing that models strong on conventional IR benchmarks still struggle on tool retrieval \cite{shi2025toolret}.
This is an important precedent for treating retrieval as a first-class agent bottleneck rather than a solved preprocessing step.
However, ToolRet focuses on tools rather than skills, and therefore does not capture the long-form procedural content, reusable prompting logic, and compositional structure of real skill libraries.
SkillFlow \cite{li2025skillflow} is complementary to our work because it proposes an agent-facing multi-stage pipeline for retrieving and selecting skills from a large community skill library, whereas \textsc{SkillRet} isolates skill retrieval as a standalone benchmark with public train/evaluation splits and controlled ranking-based evaluation.
The closest prior work is SkillRouter, which studies skill selection over roughly 80K candidate skills using a two-stage retrieve-and-rerank pipeline and a benchmark of 75 expert-verified queries \cite{zheng2026skillrouter}.
A key finding is that the full skill body carries decisive routing signal, and removing it causes large performance drops across retrieval methods \cite{zheng2026skillrouter}.
At the same time, SkillRouter is primarily a routing-model paper rather than a benchmark paper.
Its core contribution is how to design and train a scalable router, whereas our goal is to provide a broader benchmark for comparing retrieval quality across models and settings.


\section{SkillRet Benchmark}
\label{sec:skillret-benchmark}

\textsc{SkillRet} is a large-scale benchmark for retrieving relevant
agent skills from a curated library of publicly available skills. Starting
from 22{,}795 community-contributed skills, we apply quality filtering and
deduplication to obtain a curated corpus of 17{,}810 skills
(Section~\ref{sec:data-collection}). We then generate natural-language
queries that mirror realistic agent invocation patterns, where each query
requires one or more skills from the library
(Section~\ref{sec:query-generation}). Finally, we filter the generated
query--skill pairs through automatic checks, LLM-based review, an audit of
the evaluation pool for functional equivalence, and human expert validation,
yielding a benchmark of 16{,}129 skills in disjoint training and evaluation
splits. Fig.~\ref{fig:query-gen-pipeline} illustrates the full
data construction pipeline.

\subsection{Data Collection and Quality Filtering}
\label{sec:data-collection}

\paragraph{Raw corpus.}
We start from a snapshot of 22{,}795 agent skills crawled from
\texttt{claude-plugins.dev}\footnote{\url{https://claude-plugins.dev}},
a community-maintained, open-source marketplace that auto-indexes
all public agent skills on GitHub. Each record contains a skill identifier,
name, natural-language description, the full skill body (\texttt{SKILL.md}),
and marketplace metadata including GitHub stars, platform-specific install
counts, author, namespace, and license.

\paragraph{Five-stage filtering.}
We apply a pipeline to remove noise and redundancy. It has two phases.
\emph{Content eligibility} (Steps~1--3) ensures that each skill meets basic
quality and legal requirements, while \emph{deduplication} (Steps~4--5)
removes redundant entries.
\textit{(1) Description recovery and pruning.}
Listings with missing or stub descriptions ($<\,10$ characters) are recovered
via YAML frontmatter parsing or first-paragraph extraction. Unrecoverable
entries are removed (3 skills).
\textit{(2) Language filtering.}
Skills whose bodies contain more than 3\% non-Latin characters are removed (1{,}319 skills),
retaining only English-language skills.
\textit{(3) License filtering.}
Skills declaring a license other than MIT or Apache-2.0 are excluded
(255 skills). License-undeclared near-duplicates of these entries are
identified using normalized content hashes and also removed
(1{,}249 skills in total).
\textit{(4) Content deduplication.}
Each skill body is normalized by stripping YAML frontmatter, converting text
to lowercase, and removing non-alphanumeric characters, and is then hashed
using SHA-256. Among duplicate entries, we retain the entry with the highest
star and install counts (1{,}547 skills removed).
\textit{(5) Search-target deduplication.}
Skills sharing an identical normalized name--description pair are deduplicated
using a hash of the concatenated fields, again retaining the most popular
entry (867 skills removed).
Both deduplication steps rely on exact hashes and therefore cannot detect skills
that implement the same capability with different wording. We therefore apply an
LLM-based functional-equivalence audit to the evaluation pool, detailed at the
end of Section~\ref{sec:query-generation}.

After filtering, 17{,}810 skills remain (78.1\% of the raw corpus), forming
the curated document corpus from which the benchmark is drawn. The per-step
attrition is tabulated in Appendix~\ref{app:filtering-details}
(Table~\ref{tab:filtering}).
Query generation and validation, described in the following subsection, retain
only skills that are linked to at least one query, and the
functional-equivalence and capability-leakage audit removes further evaluation
skills. The final benchmark contains 16{,}129 skills, comprising a training pool
of 10{,}123 and a held-out evaluation pool of 6{,}006 paired with 4{,}392
queries.

\subsection{Skill--Query Pair Generation}
\label{sec:query-generation}

  To construct a realistic evaluation set, we generate natural-language user queries via a
  self-instruct-style~\cite{wang2023self} pipeline in which a large language model is prompted
  to produce queries that include one or more skills from the library.
  \paragraph{Seed examples.}
  To encourage lexical and structural diversity, we supply each generation call with a random
  subset of the GAIA benchmark validation set~\cite{mialon2023gaia} (165 questions) as style seeds.
  These seeds illustrate the range of tones, lengths, and request types found in realistic
  user messages, and the model is instructed to match this diversity rather than converge on
  a fixed template.

  \paragraph{Skill sampling.}
  For each generation call we sample $k \in \{1, 2, 3\}$ skills uniformly at random, where $k$
  is drawn with equal probability across the three values.
  Skills are selected via inverse-frequency weighted sampling \cite{kang2019decoupling,cui2019class}. Each skill's probability is
  proportional to $1 / (\text{freq} + 1)$, where $\text{freq}$ is the number of queries
  already generated for that skill.
  A first pass exhausts all skills with zero coverage before any skill is repeated,
  ensuring that every skill in the library is represented at least once in the final set.

  \paragraph{Query generation.}
  Each generation call receives the name and description of the sampled skills
  (without the full skill body) and is instructed to produce a single user message that
  naturally requires all selected skills.
  The prompt explicitly forbids mentioning any skill name in the generated query, forcing
  the task need to emerge from the scenario description rather than from lexical overlap with
  the skill identifier.
  Previously generated queries for the same skill are shown to the model to suppress
  near-duplicate outputs.
  Evaluation queries are generated with Claude Opus 4.6~\cite{anthropic2026claude}, while training queries are generated with Qwen3.5-122B-A10B~\cite{qwen35}.
  If the model judges the skill combination to be unrealistic, it may output a designated null token, and that combination is discarded. Full prompt details are provided in Appendix~\ref{app:query-gen-prompt}.

 \begin{figure}[t]
    \centering
    \includegraphics[width=\linewidth]{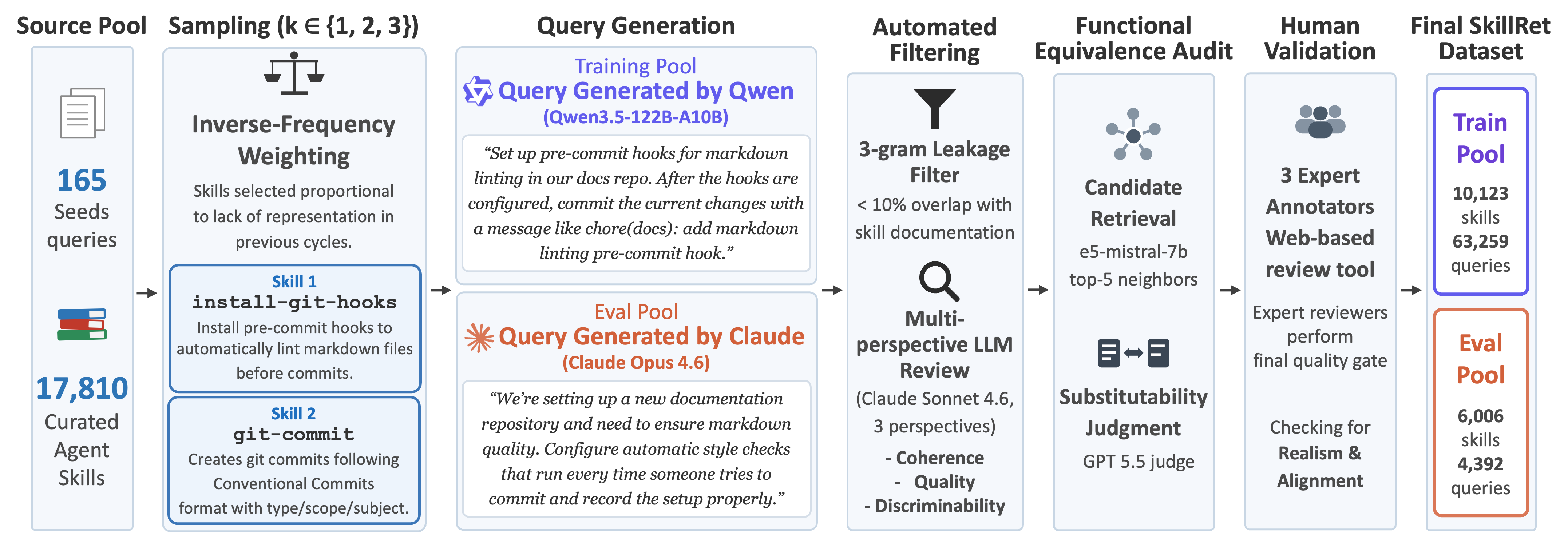}
\caption{%
Overview of the \textsc{SkillRet} data generation pipeline.
From 17{,}810 curated skills and 165 GAIA seed queries, an LLM synthesizes
user messages that require the sampled skills. Training queries use
Qwen3.5-122B-A10B and evaluation queries use Claude Opus~4.6. Automatic
filtering, LLM review, the functional-equivalence and capability-leakage
audit, and human validation yield a final evaluation set of 6{,}006 skills
and 4{,}392 queries alongside 63{,}259 training queries.
}
    \label{fig:query-gen-pipeline}
  \end{figure}

\paragraph{Automatic quality filtering.}
  Generated queries pass through a two-stage automatic filter. This procedure is
  an open rejection-sampling process in which a generator output is not accepted
  as ground truth merely because it was conditioned on the sampled skills. Failed
  candidates are discarded and regenerated.
  \textit{(1) Leakage detection.} We compute the 3-gram overlap between each query
  and its associated skill documentation. Queries whose overlap ratio exceeds a
  threshold of 10\% are flagged as leaking skill content and discarded.
  \textit{(2) Multi-perspective LLM review.} A second LLM call evaluates each
  query from three independent reviewer perspectives: skill coherence (does the
  query genuinely require the skill?), query quality (is the request specific and
  realistic?), and benchmark discriminability (would a model without the skill fail
  to answer it?). A query is rejected if two or more of the three perspectives
  return an invalid verdict. A single invalid verdict routes the query to human
  review rather than discarding it outright. Full prompts are provided in
  Appendix~\ref{app:llm-review}.

\paragraph{Functional-equivalence and capability-leakage audit.}
Exact hashes cannot detect differently worded skills that implement the same
capability. We therefore retrieve the five nearest neighbors of every evaluation
skill with e5-mistral-7b-instruct, a model outside the Qwen family used for our
retrievers. We retain candidate pairs with cosine similarity of at least 0.80 and
ask GPT-5.5 to judge functional substitutability from the complete skill bodies.
Within the evaluation pool, union--find groups substitutable pairs into
equivalence classes. We retain one representative per class, remove the redundant
entries, and remap their qrels so that no query loses a valid answer. We also
identify evaluation skills with a functional counterpart in the training pool.
Because these cross-split cases have no held-out representative, we remove the
affected evaluation skills and drop any query that relies on one rather than
shrinking its gold set. This
correction converts confirmed equivalent retrievals from false negatives into
positives and removes all detected capability-level train--evaluation overlap.
Appendix~\ref{app:label-audit} reports the full audit and graded-relevance
robustness checks.

\paragraph{Human expert validation.}
  Queries that pass automatic filtering and the audit are reviewed by three expert
  annotators using a custom web-based review tool. The tool presents each query
  alongside the associated skill name, description, and the LLM pre-judgment
  rationale, allowing annotators to assess skill--query alignment, realism, and
  discriminability. Annotators cast a binary valid/invalid mark. This stage serves
  as the final quality gate, catching subtle failures that automated filters miss,
  such as queries that are plausible in isolation but do not genuinely depend on
  the paired skill.

\paragraph{Training and evaluation splits.}
To construct query sets for both splits, we generate training queries using
Qwen3.5-122B-A10B~\cite{qwen35} and evaluation queries using Claude
Opus~4.6~\cite{anthropic2026claude}.
We deliberately use different model families for the two splits so that
retrieval models trained on the training set cannot exploit stylistic
artifacts of a single generator to inflate evaluation scores.
The larger scale of training generation (63{,}259 queries) makes the
open-weight model the practical choice. We allocate the higher-capacity
model to the evaluation set, where query quality directly affects benchmark
reliability.

The final split comprises a training pool of 10{,}123 skills and
63{,}259 queries, and an evaluation pool of 6{,}006 skills and
4{,}392 queries, with no exact skill overlap between the two sets.
As Fig.~\ref{fig:split-dist} in the Appendix shows, the major-category
distribution of each split deviates by less than 1\,percentage point from
that of the full library, confirming that the split preserves the natural
category distribution without explicit stratification.


\section{Benchmark Analysis}
\label{sec:benchmark-analysis}


\subsection{Taxonomy Overview}
\label{sec:taxonomy-overview}

We build the taxonomy over the curated corpus of 17{,}810 skills that enters
the data generation pipeline in Fig.~\ref{fig:query-gen-pipeline}, and every
skill in the final benchmark carries the resulting major and sub-category
label.
The taxonomy is constructed through a five-stage pipeline.
{(1)~Tag Discovery}: an LLM annotates each skill with three
structured tags (\emph{primary\_action}, \emph{primary\_object},
\emph{domain}) similar to \cite{gilardi2023chatgpt,ziems2024can}
{(2)~Clustering}: $k$-means over tag vectors at multiple
resolutions reveals \emph{stable clusters}, \ie groups that persist
across different values of $k$.
{(3)~Taxonomy Construction}: stable clusters seed an initial
draft, which experts iteratively refine into
{6~Major categories and 18~Sub-categories}.
{(4)~LLM-based Assignment}. Because three-axis tags capture
only surface-level attributes, we employ Claude Sonnet~4.6 to classify
all 17{,}810 skills in the curated corpus using their full name and description
(Appendix~\ref{app:llm-assignment} reports representative tag-rule
failures that motivate this design choice).
{(5)~Human Validation}. A stratified sample of 200 skills
is independently verified by experts, yielding an average
accuracy of 95.5\% for major categories and 92.2\% for
sub-categories, with full three-way agreement on 91.0\% and 84.5\%
of items respectively.
Appendix~\ref{app:taxonomy-iteration} provides full details of each
stage.
Software Engineering accounts for
62.2\% of the curated corpus while Information Retrieval comprises only 3.1\%,
mirroring the natural composition of public agent skill ecosystems
(Fig.~\ref{fig:category-dist} in the Appendix).
\begin{wraptable}{r}{0.50\textwidth}
\vspace{1.1em}
\centering
\caption{Taxonomy overview.}
\label{tab:taxonomy}
\vspace{-0.01em}
\footnotesize
\setlength{\tabcolsep}{3pt}
\resizebox{0.94\linewidth}{!}{%
\begin{tabular}{@{}llrr@{}}
\toprule
\textbf{Major Category} & \textbf{Sub-Category} & \textbf{Skills} & \textbf{\% Total} \\
\midrule
Software Eng.   & Development          & 4{,}423 & 24.8 \\
                & Analysis \& Testing  & 2{,}320 & 13.0 \\
                & Infra.\ \& DevOps    & 1{,}970 & 11.1 \\
                & Documentation        &    889 &  5.0 \\
                & Version Control      &    756 &  4.2 \\
                & Security             &    727 &  4.1 \\
\midrule
AI Agents       & Agent Development    & 1{,}194 &  6.7 \\
                & Agent Orchestration  &    607 &  3.4 \\
                & Agent Evaluation     &    273 &  1.5 \\
\midrule
Business        & Business Analysis    &    821 &  4.6 \\
\& Planning     & Project Mgmt.        &    788 &  4.4 \\
\midrule
Data \& ML      & ML Development       &    477 &  2.7 \\
                & Data Engineering     &    418 &  2.3 \\
                & Data Analysis        &    416 &  2.3 \\
\midrule
Content         & Writing \& Text      &    687 &  3.9 \\
Creation        & Visual \& Media      &    489 &  2.7 \\
\midrule
Info.\          & General Search       &    357 &  2.0 \\
Retrieval       & Technical Search     &    198 &  1.1 \\
\bottomrule
\end{tabular}%
}
\vspace{-1.0em}
\end{wraptable}

\subsection{Skill \& Taxonomy Statistics}
\label{sec:skill-taxonomy-stats}
\begin{figure}[t]
\centering
\hspace{-6mm}
\includegraphics[width=1.03\linewidth]{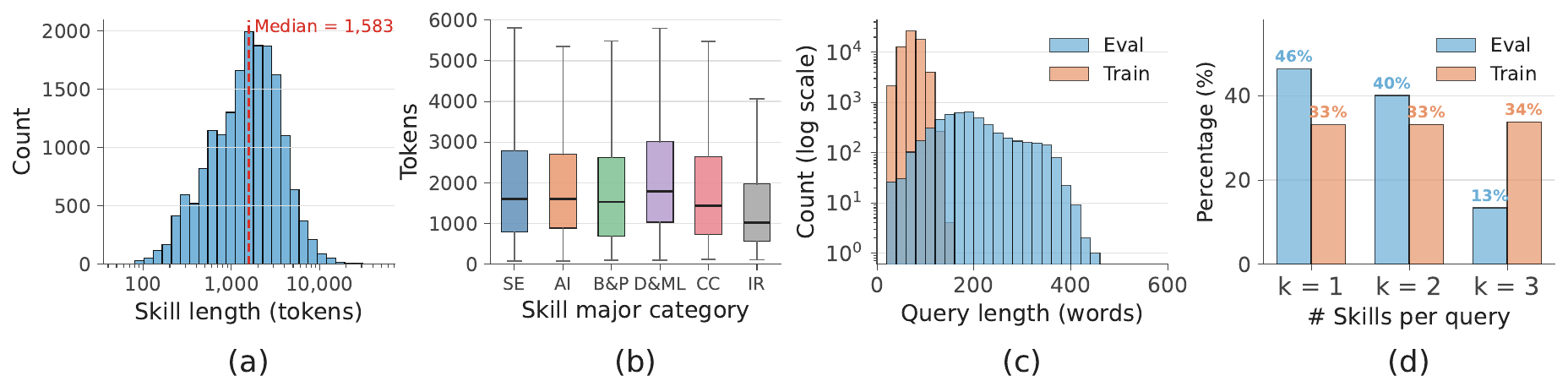}
\vspace{-2mm}
\caption{Skill and query length statistics.
(a) Distribution of document length across the curated corpus of 17{,}810
skills prior to the data generation pipeline.
(b) Box plots of document length by major category.
(c) Query length distributions for the pre-audit evaluation set and training set.
(d) Distribution of $k$ (number of skills per query) in each split; training queries are sampled uniformly across $k$, whereas evaluation queries are concentrated on $k{=}1$ and $k{=}2$. }
\label{fig:skill-query-stats}
\end{figure}

Each skill is represented as the composite text
\texttt{name\,|\,description\,|\,skill\_md}, which is the
actual retrieval target used by all models in our evaluation.
The \texttt{skill\_md} component contains the full Markdown body
including instructions, decision logic, usage constraints, and
implementation details.
Measured in \texttt{cl100k\_base} tokens, this composite text has a
{median length of 1{,}583} tokens
(mean~2{,}083; 95th~percentile~5{,}531; max~47{,}412),
resulting in approximately 37.1\,M tokens across the corpus
(Fig.~\ref{fig:skill-query-stats}\,(a)).
This is an order of magnitude longer than typical tool descriptions
in existing benchmarks~\cite{shi2025toolret}, making skill retrieval
a fundamentally long-document matching problem.
Fig.~\ref{fig:skill-query-stats}\,(b) shows the per-Major length
distributions. Data~\&~ML skills are the longest
(median~1{,}795 tokens). Information Retrieval skills
are the shortest, reflecting their comparatively concise search-oriented instructions.

\subsection{Query Statistics}

We further summarize the key distributional properties of the generated queries across the pre-audit training and evaluation splits. In Fig.~\ref{fig:skill-query-stats}(c), evaluation queries, generated by Claude Opus~4.6, are substantially longer than training queries generated by Qwen3.5-122B-A10B \cite{qwen35}, with a median length of 170 words versus 72 words and a 95th percentile of 270 versus 108 words. This difference likely reflects generation style across the two model families, where Opus~4.6 tends to produce more detailed, scenario-rich requests, making evaluation queries inherently more challenging for lexical matching methods. In terms of the number of required skills per query, training queries are distributed uniformly across $k \in \{1,2,3\}$, whereas evaluation queries are concentrated on lower values of $k$, with 46\% single-skill queries, 40\% two-skill queries, and 13\% three-skill queries (Fig.~\ref{fig:skill-query-stats}(d)). Notably, multi-skill queries ($k \geq 2$) account for the majority of the pre-audit evaluation set (54\%), motivating the corrected-set analysis in Section~\ref{sec:results}.

\begin{table*}[t]
  \centering
  \caption{First-stage retrieval results on the corrected \textsc{SkillRet} evaluation set (6,006 skills; 4,392 queries). Functional equivalents are remapped and train--evaluation functional overlaps are removed. Best result per metric is \textbf{bolded}.}
  \label{tab:main-results}
  \vspace{0.4em}
  \small
  \resizebox{\textwidth}{!}{%
  \begin{tabular}{llc ccc ccc ccc}
  \toprule
  \multirow{2}{*}{\textbf{Type}} & \multirow{2}{*}{\textbf{Model}} & \multirow{2}{*}{\textbf{Params}}
  & \multicolumn{3}{c}{\textbf{NDCG}} & \multicolumn{3}{c}{\textbf{Recall}} & \multicolumn{3}{c}{\textbf{Completeness}} \\
  \cmidrule(lr){4-6} \cmidrule(lr){7-9} \cmidrule(lr){10-12}
  & & & @5 & @10 & @15 & @5 & @10 & @15 & @5 & @10 & @15 \\
  \midrule
  Sparse & BM25~\cite{robertson1994some} & -- & 49.31 & 51.69 & 52.75 & 53.03 & 59.41 & 62.92 & 38.21 & 44.56 & 47.93 \\
  \midrule
  \multirow{5}{*}{Encoder}
  & e5-small-v2~\cite{wang2022text} & 118M & 42.63 & 44.66 & 45.66 & 46.03 & 51.59 & 54.89 & 32.61 & 37.45 & 40.35 \\
  & e5-large-v2~\cite{wang2022text} & 335M & 51.30 & 53.41 & 54.44 & 54.88 & 60.55 & 63.89 & 40.16 & 45.45 & 48.68 \\
  & bge-small-en-v1.5~\cite{xiao2024c} & 33M & 52.57 & 54.51 & 55.45 & 54.73 & 60.01 & 63.07 & 38.96 & 43.97 & 47.11 \\
  & snowflake-arctic-embed-s~\cite{merrick2024arctic} & 33M & 54.48 & 56.39 & 57.35 & 56.95 & 61.96 & 65.02 & 41.37 & 46.22 & 49.32 \\
  & bge-large-en-v1.5~\cite{xiao2024c} & 335M & 57.04 & 59.00 & 59.80 & 59.19 & 64.37 & 66.95 & 42.96 & 48.34 & 50.80 \\
  \midrule
  \multirow{13}{*}{Decoder}
  & F2LLM-v2-80M~\cite{zhang2026f2llm} & 80M & 46.26 & 48.22 & 49.27 & 49.51 & 54.86 & 58.32 & 34.77 & 39.82 & 43.06 \\
  & pplx-embed-v1-0.6b~\cite{eslami2026diffusion} & 0.6B & 51.71 & 54.26 & 55.52 & 56.72 & 63.60 & 67.70 & 41.51 & 48.27 & 52.32 \\
  & NV-Embed-v1~\cite{lee2024nv} & 7B & 54.62 & 57.03 & 57.96 & 58.16 & 64.64 & 67.64 & 42.12 & 48.00 & 50.87 \\
  & KaLM-Gemma3-12B~\cite{zhao2025kalm} & 12B & 56.31 & 58.95 & 60.11 & 60.48 & 67.56 & 71.35 & 44.49 & 52.37 & 56.42 \\
  & Qwen3-Embedding-0.6B~\cite{zhang2025qwen3} & 0.6B & 59.92 & 61.94 & 62.98 & 62.59 & 67.87 & 71.25 & 45.88 & 51.09 & 54.67 \\
  & jina-embeddings-v5-text-small~\cite{akram2026jina} & 0.6B & 60.85 & 62.96 & 63.90 & 63.46 & 68.99 & 72.02 & 46.93 & 53.26 & 56.74 \\
  & Qwen3-Embedding-8B~\cite{zhang2025qwen3} & 8B & 61.35 & 63.64 & 64.73 & 64.24 & 70.29 & 73.76 & 47.43 & 54.14 & 58.15 \\
  & harrier-oss-v1-270m~\cite{huang2026harrier} & 270M & 62.42 & 64.56 & 65.33 & 65.12 & 70.80 & 73.27 & 49.00 & 55.37 & 58.06 \\
  & Octen-Embedding-8B~\cite{octen2025rteb} & 8B & 64.00 & 65.93 & 66.85 & 65.98 & 71.07 & 73.99 & 49.32 & 55.35 & 58.99 \\
  & harrier-oss-v1-0.6b~\cite{huang2026harrier} & 0.6B & 68.16 & 70.26 & 71.09 & 70.64 & 76.16 & 78.76 & 54.80 & 61.61 & 64.82 \\
  & SkillRouter-Embedding-0.6B~\cite{zheng2026skillrouter} & 0.6B & 71.80 & 73.54 & 74.33 & 73.73 & 78.18 & 80.68 & 57.22 & 62.70 & 65.96 \\
  & SkillRet-Embedding-0.6B~(ours) & 0.6B & 79.02 & 81.12 & 81.87 & 82.38 & 87.74 & 90.16 & 70.83 & 78.94 & 82.65 \\
  & SkillRet-Embedding-8B~(ours) & 8B & \textbf{84.58} & \textbf{86.44} & \textbf{86.95} & \textbf{88.55} & \textbf{93.25} & \textbf{94.84} & \textbf{80.35} & \textbf{88.11} & \textbf{90.80} \\
  \bottomrule
  \end{tabular}%
  }
\end{table*}

\section{Evaluation}
\label{sec:evaluation}

\subsection{Experimental Setup}
\label{sec:exp-setup}

\paragraph{Setup.}
We adopt a two-stage retrieve-then-rerank pipeline where an embedding model retrieves
the top-$k$ candidates via cosine similarity and a reranker re-scores each
query--candidate pair.
Larger $k$ yields better coverage but increases reranking cost.
We evaluate $k \in \{10, 20, 50\}$ and set $k{=}20$ considering the trade-off
between retrieval quality and computational cost.
Ablation results are in Appendix~\ref{app:topk-ablation}.
Encoding the full document text, including the name, description, and Markdown body,
consistently outperforms encoding name and description only,
as shown in Appendix~\ref{app:doc-repr-ablation}.
We therefore encode each document up to the model's maximum sequence length
for all experiments, with per-model limits listed in Appendix~\ref{app:max-seq-lengths}.
We use each model's officially recommended prompts.
Only the Harrier, Qwen3-Embedding, and Qwen3-Reranker families have their default
web-search instruction replaced with a skill-retrieval instruction we authored. 
Full specifications are in Appendix~\ref{app:prompts-eval}.

\paragraph{Models.}
For first-stage retrieval, we evaluate 19 systems across three categories, including the sparse BM25 baseline, encoder-only models, and decoder-only models, covering sub-100M to 12B parameters.
The full list is in Table~\ref{tab:main-results}.
For reranking, we evaluate jina-reranker-v2-base-multilingual~\cite{jinaai2024rerankerv2} and the
Qwen3-Reranker~\cite{zhang2025qwen3} family at scales 0.6B, 4B, and 8B.
We also include SkillRouter-Embedding-0.6B~\cite{zheng2026skillrouter} and
SkillRouter-Reranker-0.6B~\cite{zheng2026skillrouter} as external fine-tuned baselines,
evaluated using the publicly released checkpoints on HuggingFace.
We refer to all models fine-tuned on \textsc{SkillRet} training data collectively
as the \textsc{SkillRet} model family, comprising
SkillRet-Embedding-0.6B, SkillRet-Embedding-8B,
and SkillRet-Reranker-0.6B.
Although Harrier-OSS outperforms Qwen3-Embedding off-the-shelf,
we use Qwen3-Embedding as our fine-tuning base because Harrier is itself
a fine-tuned derivative of Qwen3-Embedding.
We verify that fine-tuning from either base yields comparable results in Appendix~\ref{app:base-model-selection}.

\paragraph{Training details.}
We fine-tune Qwen3-Embedding-0.6B and Qwen3-Embedding-8B using
MultipleNegativesRankingLoss with in-batch negatives
on 127{,}190 positive query--skill pairs derived from the training split.
SkillRet-Reranker-0.6B is fine-tuned from Qwen3-Reranker-0.6B using binary cross-entropy
on the \texttt{yes}/\texttt{no} token probability at the final decoding position.
Hard negatives are mined with the fine-tuned SkillRet-Embedding-0.6B retriever.
Full hyperparameters are in Appendix~\ref{app:finetuning-details}.

\paragraph{Evaluation metrics.}
We report three primary metrics at $k{=}10$.
\textbf{NDCG@10}~\cite{jarvelin2002cumulated} measures ranking quality,
\textbf{Recall@10} measures the fraction of ground-truth skills retrieved,
and \textbf{Completeness@10}~\cite{qu2024towards} measures the fraction of queries
where all ground-truth skills are retrieved, i.e., Recall@10${=}1$.
All evaluations are run on a single NVIDIA B200 GPU (180\,GB VRAM) per model
to ensure reproducibility.

\subsection{Experimental Results}
\label{sec:results}

\paragraph{Embedding Retrieval.}
Table~\ref{tab:main-results} reports the retrieval performance of all evaluated
models on the corrected \textsc{SkillRet} benchmark.
The best encoder-only model, bge-large-en-v1.5~\cite{xiao2024c}, reaches 59.00
NDCG@10, while the strongest off-the-shelf model,
harrier-oss-v1-0.6b~\cite{huang2026harrier}, reaches 70.26.
Decoder-only models support maximum sequence lengths of 8K--32K tokens,
far exceeding the 512-token limit of encoder-only models,
and can thus encode full skill documents without truncation.
Model scale alone is insufficient. NV-Embed-v1~\cite{lee2024nv} at 7B scores
57.03, below the 270M Harrier model at 64.56.
What matters more is exposure to domain-relevant training data.
SkillRouter-Embedding-0.6B~\cite{zheng2026skillrouter} reaches 73.54, while
SkillRet-Embedding-0.6B and 8B reach 81.12 and 86.44.
Thus, the 8B model leads the strongest prior retriever by 12.9 points and the
strongest off-the-shelf retriever by 16.2 points.
The functional-equivalence correction raises all 19 models by a mean of 3.29
points (range 2.70--3.91) with zero inversions among 171 model pairs, showing
that false negatives affected absolute scores but did not explain the
fine-tuning gain or model ordering.

\paragraph{Reranking.}
Table~\ref{tab:rerank-results} separates first-stage recall from reranker
quality. In the +Rerank setting, the reranker rescores the 20 candidates
returned by the first stage. In +Oracle, every gold skill missing from that pool
is first inserted by displacing an equal number of lowest-ranked non-gold
candidates, holding the pool at 20, and the same reranker then rescores it.
Starting from SkillRet-Embedding-8B at 86.44 NDCG@10, SkillRet-Reranker reaches
87.74. The oracle pool raises this only to 88.89.
SkillRouter-Reranker~\cite{zheng2026skillrouter} reaches 86.78 on the same
candidate pool, and its corresponding gap is 1.27 points. These small oracle
gains show that the strongest first stage has nearly exhausted top-20 recall.
The remaining ceiling is primarily reranker ability rather than candidate
coverage.

\begin{table}[t]
    \centering
    \caption{Oracle top-20 analysis on the corrected evaluation set. ``+Oracle'' inserts missing gold skills before applying the same reranker.}
    \label{tab:rerank-results}
    \small
    \begin{tabular}{llccc}
    \toprule
    \textbf{Reranker} & \textbf{Setting} & \textbf{NDCG@5} & \textbf{NDCG@10} & \textbf{NDCG@15} \\
    \midrule
    \multirow{3}{*}{SkillRouter-Reranker-0.6B~\cite{zheng2026skillrouter}}
        & +Rerank & 84.69 & 86.78 & 87.33 \\
        & +Oracle & 85.57 & 88.05 & 88.92 \\
        & Gain    & +0.88 & +1.27 & +1.59 \\
    \midrule
    \multirow{3}{*}{SkillRet-Reranker-0.6B~(ours)}
        & +Rerank & 86.10 & 87.74 & 88.21 \\
        & +Oracle & 86.77 & 88.89 & 89.63 \\
        & Gain    & +0.66 & +1.14 & +1.42 \\
    \bottomrule
    \end{tabular}
\end{table}

\subsection{Analysis}

\paragraph{Multi-skill queries amplify the fine-tuning gain.}
Table~\ref{tab:multi-skill-main} stratifies the corrected evaluation set by the
number $k$ of required skills. Off-the-shelf Qwen3-Embedding-0.6B degrades
sharply as $k$ grows, completing only 1.8\% of three-skill queries.
SkillRet-Embedding-0.6B remains substantially more robust. Its NDCG@10 advantage
grows from 10.3 points at $k{=}1$ to 33.1 points at $k{=}3$, and its three-skill
completeness reaches 42.1\%. Appendix~\ref{app:multi-skill} breaks this trend down by major category.

\begin{table}[t]
\centering
\caption{Performance by number of required skills on the corrected evaluation set.}
\label{tab:multi-skill-main}
\small
\setlength{\tabcolsep}{3.5pt}
\begin{tabular}{lccc|ccc}
\toprule
& \multicolumn{3}{c|}{\textbf{Completeness@10}} & \multicolumn{3}{c}{\textbf{NDCG@10}} \\
\textbf{Model} & $k{=}1$ & $k{=}2$ & $k{=}3$ & $k{=}1$ & $k{=}2$ & $k{=}3$ \\
\midrule
Qwen3-Emb.-0.6B    & 83.8 & 25.8 & 1.8 & 74.0 & 53.6 & 40.6 \\
SkillRet-Emb.-0.6B & \textbf{92.1} & \textbf{74.2} & \textbf{42.1} & \textbf{84.2} & \textbf{79.6} & \textbf{73.7} \\
\bottomrule
\end{tabular}
\end{table}

\paragraph{Deployment efficiency.}
We measure 100 fixed queries with warm kernels on one NVIDIA B200 using HuggingFace Transformers, FlashAttention-2, bfloat16, and each architecture's standard implementation. The 0.6B and 8B embedding models require 21.25 and 26.25\,ms per query, respectively, whereas reranking 20 candidates costs 831\,ms at 0.6B and 1.88\,s at 8B. Thus, first-stage retrieval is real-time and reranking is the principal bottleneck. Given that the 8B embedder alone reaches 86.44 NDCG@10 and reranking adds only 1.30 points, latency-sensitive deployments can omit the second stage. Full latency and throughput results are in Appendix~\ref{app:efficiency}.

\paragraph{End-to-end agent utility.}
We integrate SkillRet-Embedding-0.6B into a Claude Code agent with Claude Opus~4.6 and evaluate 80 Terminal-Bench tasks~\cite{merrill2026terminalbench} over five independent trials. A \texttt{UserPromptSubmit} hook retrieves from the 16{,}783 skills available
before the functional-equivalence audit and injects the top three skill bodies
when cosine similarity is at least 0.30. As Table~\ref{tab:terminal-bench} shows, retrieval matches the no-retrieval baseline within 0.3 percentage points while reducing mean cost per trial by 10\%. Because Pass@1 depends only on task completion and has no gold-skill label, this result also avoids the benchmark's false-negative issue. Any functionally substitutable skill that helps solve the task receives full credit.

\paragraph{Training effect: skill-relevant sentence focus.}

Fine-tuned models substantially outperform their base counterparts,
as shown in Table~\ref{tab:main-results}, but why does training help?
We hypothesize that fine-tuning does not simply improve overall query
encoding, but instead sharpens the model's focus on the small subset of
sentences within a query that directly signals skill intent.
Skill queries are typically long, scenario-rich requests in which only
a few sentences carry the actionable capability signal.
The remainder consists of background context, output requirements, and constraints
largely orthogonal to skill selection.
A base model may distribute attention broadly across all sentences,
while a fine-tuned model learns via retrieval supervision to prioritize
the sentences that most directly determine which skill is needed.

\begin{table*}[t]
\centering

\begin{minipage}[t]{0.48\textwidth}
\centering
\caption{End-to-end evaluation on 80 Terminal-Bench tasks over five trials.
Both conditions use Claude Code with Claude Opus~4.6.}
\label{tab:terminal-bench}
\end{minipage}
\hfill
\begin{minipage}[t]{0.48\textwidth}
\centering
\caption{Effect of masking important query snippets on retrieval performance
(NDCG@10).}
\label{tab:query-mask-ablation-main}
\end{minipage}

\vspace{0.5em}

\begin{minipage}[t]{0.48\textwidth}
\vspace{0pt}
\centering
\small
\renewcommand{\arraystretch}{1.67}
\setlength{\tabcolsep}{6pt}
\begin{tabular}{lcc}
\toprule
\textbf{Skill retrieval} & \textbf{Pass@1} & \textbf{Cost/trial} \\
\midrule
None            & 65.5\% & \$0.86 \\
SkillRet (ours) & 65.8\% & \$0.78 \\
\bottomrule
\end{tabular}
\end{minipage}
\hfill
\begin{minipage}[t]{0.48\textwidth}
\vspace{0pt}
\centering
\small
\renewcommand{\arraystretch}{1.00}
\setlength{\tabcolsep}{5pt}
\begin{tabular}{lcc}
\toprule
\textbf{Condition} & \textbf{Base} & \textbf{Trained} \\
\midrule
Full query   & 73.41           & 80.89           \\
mask\_top\_1 & 56.34 (-23.3\%) & 57.29 (-29.2\%) \\
mask\_top\_2 & 40.39 (-45.0\%) & 38.92 (-51.9\%) \\
mask\_top\_3 & 26.66 (-63.7\%) & 24.00 (-70.3\%) \\
\bottomrule
\end{tabular}
\end{minipage}

\end{table*}

To test this, we conduct a sentence erasure analysis~\cite{barkan2024llm,li2016understanding}
on the 2,319 single-skill queries from the pre-audit evaluation split.
For each sentence $s_i$ in query $q$, we replace it with \texttt{[MASK]},
re-encode the masked query $q_{\setminus s_i}$, and compute
$\mathrm{importance}(s_i)=\mathrm{sim}(q,d^{+})-\mathrm{sim}(q_{\setminus s_i},d^{+})$.
We then mask the top-$k$ most important sentences and re-run retrieval,
with results shown in Table~\ref{tab:query-mask-ablation-main}.
On the full query, the trained model outperforms the base model by 7.5 NDCG@10 points,
yet removing the single most important sentence causes a larger performance drop.
This suggests that the trained model concentrates its retrieval signal on a small set of skill-relevant sentences, whereas the base model relies more diffusely on information spread across the entire query. A qualitative visualization of this pattern is shown in Fig.~\ref{fig:loo-ex} in Appendix~\ref{app:loo-viz}.

\paragraph{MTEB Retrieval ranking does not predict skill retrieval performance.}
\label{sec:mteb-comparison}

\begin{figure}[H]
\centering

\begin{minipage}[t]{0.53\textwidth}
\vspace{0pt}
\centering
\includegraphics[width=\linewidth]{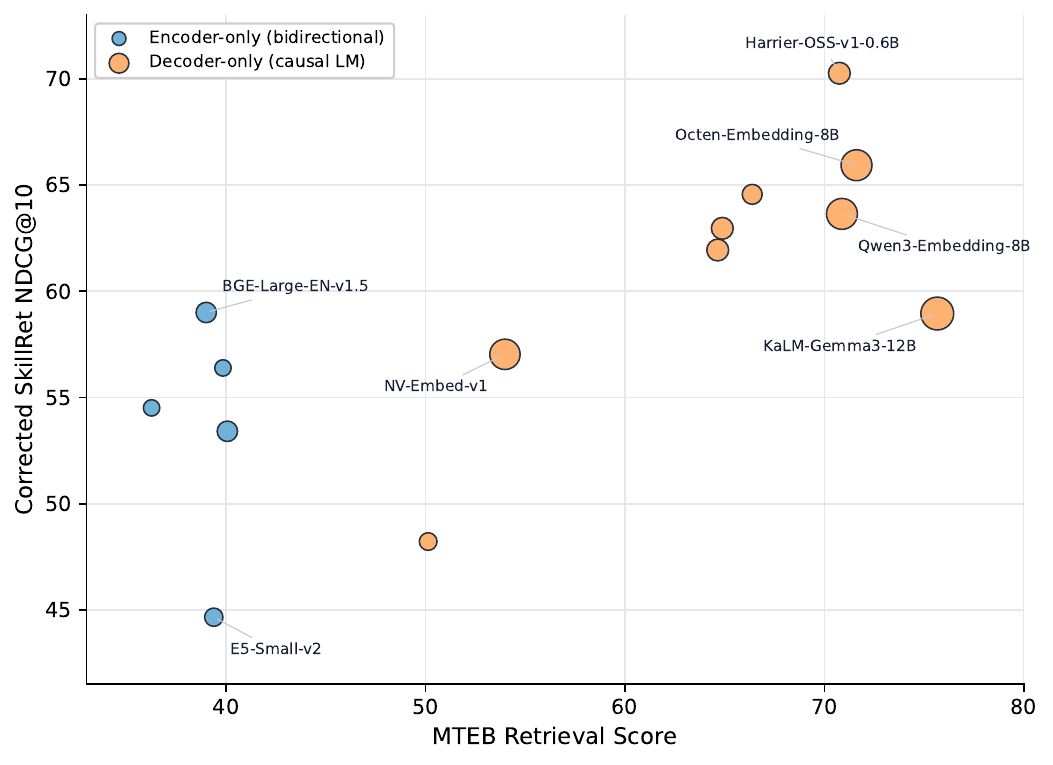}
\vspace{-2.0em}
\captionof{figure}{MTEB Retrieval score vs.\ \textsc{SkillRet}.
Circle size is proportional to parameter count.}
\label{fig:mteb-scatter}
\end{minipage}
\hfill
\begin{minipage}[t]{0.45\textwidth}
\vspace{0pt}
\centering
\captionsetup{font=footnotesize}
\captionof{table}{Per-major-category NDCG@10 on the corrected evaluation set for Qwen3-Embedding (Base) and SkillRet-Embedding (Ours), hardest first. A query is counted in every major category among its gold skills.}
\label{tab:per-category}
\vspace{0.3em}
\fontsize{8}{7.2}\selectfont
\renewcommand{\arraystretch}{1.15}
\setlength{\tabcolsep}{1.8pt}
\begin{tabular}{@{}l rr rr rr@{}}
\toprule
& \multicolumn{2}{c}{\textbf{0.6B}} & \multicolumn{2}{c}{\textbf{8B}} & \multicolumn{2}{c}{\textbf{$\Delta$}} \\
\cmidrule(lr){2-3} \cmidrule(lr){4-5} \cmidrule(lr){6-7}
\textbf{Major} & Base & Ours & Base & Ours & 0.6B & 8B \\
\midrule
Info.\ Retr.   & 44.1 & 72.8 & 42.8 & 79.0 & +28.7 & +36.1 \\
AI Agents      & 46.6 & 73.0 & 47.5 & 81.2 & +26.5 & +33.8 \\
Bus.\ \& Plan. & 54.2 & 79.7 & 55.2 & 84.9 & +25.4 & +29.7 \\
Content Cr.    & 55.6 & 82.7 & 56.6 & 87.7 & +27.1 & +31.1 \\
Software Eng.  & 62.1 & 81.2 & 64.7 & 86.9 & +19.1 & +22.2 \\
Data \& ML     & 67.9 & 85.7 & 72.0 & 90.4 & +17.8 & +18.4 \\
\midrule
Overall        & 61.9 & 81.1 & 63.6 & 86.4 & +19.2 & +22.8 \\
\bottomrule
\end{tabular}
\end{minipage}

\vspace{-0.8em}
\end{figure}

Figure~\ref{fig:mteb-scatter} plots MTEB Retrieval score~\cite{mteb-leaderboard} against \textsc{SkillRet} NDCG@10.
MTEB Retrieval score shows a moderate positive correlation at Spearman $\rho = 0.70$,
yet ranking inversions are common, with models that score highly on MTEB often underperforming on \textsc{SkillRet}, and vice versa.
These inversions suggest that skill retrieval demands a form of query understanding
distinct from general semantic matching, requiring models to identify specific capability
signals within long, multi-sentence queries.
Task-specific fine-tuning, as demonstrated by the SkillRet model family, is the most
effective way to bridge this gap.
Full scores and detailed examples are in Appendix~\ref{app:mteb-comparison}.

\paragraph{Per-category performance.}
\label{sec:per-category}
Table~\ref{tab:per-category} breaks down NDCG@10 on the corrected evaluation set
by the six major categories. The SkillRet models improve over the
Qwen3-Embedding baselines in every category, with gains ranging from 18.4 to
36.1 points for the 8B variant. Despite these gains the difficulty ordering is
stable across all four configurations. Information Retrieval and AI Agents
consistently score lowest, and an 11.4 point gap between the easiest and hardest
categories persists even for SkillRet-Embedding-8B. This category-level
disparity is invisible to the aggregate NDCG@10 of 86.4 and can only be surfaced
through the taxonomy-based stratification. Finer-grained sub-category results in
Appendix~\ref{app:per-sub-category} expose within-major variance of up to 11.5
points, further confirming the taxonomy's value as a diagnostic tool for
pinpointing retrieval bottlenecks.

\vspace{-2mm}

\section{Limitations}
\vspace{-3mm}
\textsc{SkillRet} has two main limitations. First, \textsc{SkillRet}
queries are designed to resemble realistic user requests but are
synthetically generated rather than collected from live agent interactions.
Thus, the evaluation set may under-represent terse, underspecified,
conversational, or user-context-dependent requests common in real
deployments. We mitigate this with GAIA-style seed examples, skill-name
leakage filtering, and query--skill validation, but bridging synthetic
benchmarks with real agent traffic remains important future work.
Second, \textsc{SkillRet} evaluates retrieval quality in isolation and does
not measure downstream task success or end-to-end agent performance.
Higher NDCG@10 does not necessarily imply better skill use, since retrieved
skills must still be selected, composed, interpreted, and executed under
practical context and latency constraints. We leave the joint study of
skill retrieval and downstream execution to future work.

\section{Conclusion}
\vspace{-3mm}
We introduced \textsc{SkillRet}, a large-scale benchmark for skill retrieval
in LLM agents, built from 16{,}129 curated skills with a two-level taxonomy
of 6 major and 18 sub-categories, a corrected evaluation set of 6{,}006
skills and 4{,}392 queries, and a training pool of 63{,}259 queries.
Unlike prior tool retrieval benchmarks, \textsc{SkillRet} targets long-form,
compositional skill documents, where the relevant signal must be matched
against a small actionable portion of the user query. Across various
embedding models, the strongest off-the-shelf retriever reaches 70.26
NDCG@10, while the strongest prior skill-retrieval model reaches 73.54.
Domain-specific fine-tuning on \textsc{SkillRet} lifts NDCG@10 to 86.44,
corresponding to a +12.9-point gain over the strongest prior retriever and
a +16.2-point gain over the strongest off-the-shelf retriever. These results
position skill retrieval as a distinct long-document matching problem and
establish \textsc{SkillRet} as a foundation for retrieval-oriented training
and benchmarking in future agent systems.

\section*{Acknowledgments}
We sincerely thank Hyojung Han and Seunghun Jeon for their helpful discussions during the early stages of this project.

\bibliographystyle{plain}
\bibliography{main}

\clearpage
\section*{Appendix}


\appendix

\section{Data, code, and model.}
We release \textsc{SkillRet} publicly at
\url{https://huggingface.co/datasets/ThakiCloud/SKILLRET}.
Code is available at
\url{https://github.com/ThakiCloud/SKILLRET}.
Pre-trained model weights are also publicly released:
SkillRet-Embedding-0.6B at
\url{https://huggingface.co/ThakiCloud/SKILLRET-Embedding-0.6B}
and SkillRet-Embedding-8B at
\url{https://huggingface.co/ThakiCloud/SKILLRET-Embedding-8B}.
The dataset is derived from public GitHub-hosted skills and synthetic queries;
it is intended for skill retrieval evaluation and model development, not for
profiling individual authors, inferring personal attributes, or certifying
end-to-end agent safety.

\section{Dataset Construction Details}
\label{app:dataset-construction}

This appendix provides supporting details for the SkillRet skill library
summarized in Section~\ref{sec:skillret-benchmark} and the taxonomy
presented in Section~\ref{sec:benchmark-analysis}:
the per-step filtering attrition (\S\ref{app:filtering-details}),
the two-pass LLM tagging procedure (\S\ref{app:tagging}),
consensus clustering over action--object combinations (\S\ref{app:clustering}),
the iterative taxonomy design process (\S\ref{app:taxonomy-iteration}),
LLM-based skill assignment (\S\ref{app:llm-assignment}),
and human validation (\S\ref{app:human-validation}).

\subsection{Per-Step Filtering Attrition}
\label{app:filtering-details}

Table~\ref{tab:filtering} reports the per-step attrition of the five-stage
filtering pipeline described in \S\ref{sec:data-collection}. The two largest
reductions come from content deduplication (Step~4) and language filtering
(Step~2).

\begin{table}[h!]
\centering
\caption{Quality filtering pipeline.
The largest reductions come from content deduplication (Step~4)
and language filtering (Step~2).}
\label{tab:filtering}
\small
\begin{tabular}{cl rr}
\toprule
\textbf{Step} & \textbf{Filter} & \textbf{Removed} & \textbf{Remaining} \\
\midrule
-- & Raw corpus & -- & 22{,}795 \\
\multicolumn{4}{l}{\emph{Phase 1: Content eligibility}} \\
1 & Description recovery / pruning & 3 & 22{,}792 \\
2 & Language filter ({$>$}\,3\% non-Latin) & 1{,}319 & 21{,}473 \\
3 & License filter + propagated cleanup & 1{,}249 & 20{,}224 \\
\multicolumn{4}{l}{\emph{Phase 2: Deduplication}} \\
4 & Content deduplication (normalized hash) & 1{,}547 & 18{,}677 \\
5 & Search-target deduplication (name+desc) & 867 & \textbf{17{,}810} \\
\bottomrule
\end{tabular}
\end{table}

\subsection{Structured Tagging}
\label{app:tagging}

To characterize each skill along interpretable dimensions, we assign three
structured tags per skill: \textbf{primary\_action} (what the skill
\emph{does}), \textbf{primary\_object} (what the skill \emph{acts on}),
and \textbf{domain} (the technical field it belongs to). We use a two-pass
procedure with Claude Sonnet~4.6.

\subsubsection{Pass 1: Category Discovery}
\label{app:pass1}

All 17{,}810 skill names with truncated descriptions (100 characters) are
submitted in a single prompt. The model is instructed to discover natural,
non-overlapping categories for each dimension at an appropriate granularity
(roughly 8--15 categories). This yields 13 actions, 14 objects, and 13 domains.
The categories are discovered by the LLM from the corpus rather than being
predefined by the authors, though the target granularity (8--15 per
dimension) is specified in the prompt. The resulting label sets were
manually reviewed by the authors to verify semantic coherence and adjust
ambiguous or overlapping categories.

\paragraph{System prompt.}
\begin{quote}
\small
\begin{verbatim}
You are a skill taxonomy analyst. You will receive a list
of ~17,000 AI coding skill names with short descriptions.

Your task: analyze ALL skills and discover the natural
categories that exist across three dimensions.

For each dimension, identify **distinct, non-overlapping
categories** at an appropriate granularity level (roughly
8-15 categories per dimension). Each category should have
a short lowercase label (1-2 words, snake_case) and a
brief description.

Dimensions:
1. **primary_action**: What the skill DOES
   (the core verb/activity)
2. **primary_object**: What the skill acts ON
   (the target/subject)
3. **domain**: What technical field the skill belongs to

Output strict JSON with this structure:
{
  "primary_action": [
    {"label": "...", "description": "..."}
  ],
  "primary_object": [
    {"label": "...", "description": "..."}
  ],
  "domain": [
    {"label": "...", "description": "..."}
  ]
}

No markdown fences, no explanations outside the JSON.
\end{verbatim}
\end{quote}

\paragraph{User message format.}
\begin{quote}
\small
\begin{verbatim}
Here are all the skills:

{skill_name_1}: {description_first_100_chars}
{skill_name_2}: {description_first_100_chars}
...
{skill_name_17810}: {description_first_100_chars}
\end{verbatim}
\end{quote}

\subsubsection{Pass 2: Batch Classification}
\label{app:pass2}

The discovered categories are injected into the system prompt as a closed
label set. Skills are then classified in batches of 100, with the model
selecting exactly one label per dimension for each skill. The output is a
structured JSON record per skill. After deduplication of any double-tagged
entries, we obtain a clean set of 17{,}810 (id, action, object, domain)
tuples.

\paragraph{System prompt.}
\begin{quote}
\small
\begin{verbatim}
You are a skill taxonomy classifier. For each AI coding
skill, assign exactly 3 labels.

**primary_action** -- choose ONE from:
  - implement: Writing, building, or creating new code,
    features, components, or systems
  - debug: Finding, diagnosing, and fixing bugs, errors,
    or unexpected behavior
  - review: Evaluating, auditing, or assessing code,
    documentation, or designs for quality
  - test: Writing, running, or managing automated tests
    and test strategies
  - design: Architecting systems, designing APIs,
    planning schemas, or defining specifications
  - document: Creating, updating, or generating
    documentation, comments, or explanations
  - refactor: Restructuring or improving existing code
    without changing behavior
  - configure: Setting up, installing, or configuring
    tools, environments, or services
  - deploy: Building, packaging, releasing, or deploying
    software to environments
  - analyze: Investigating, researching, profiling, or
    extracting insights from code or data
  - generate: Producing artifacts like images, content,
    reports, or boilerplate automatically
  - orchestrate: Coordinating, routing, or managing
    multiple agents, tasks, or workflows
  - search: Finding, discovering, or retrieving
    information from code, docs, or the web

**primary_object** -- choose ONE from:
  - code: Source code files, functions, classes, modules
  - api: REST, GraphQL, gRPC interfaces and endpoints
  - database: Database schemas, queries, migrations
  - ui_component: Frontend components, pages, layouts
  - test_suite: Unit tests, integration tests, E2E tests
  - documentation: READMEs, API docs, guides, changelogs
  - pipeline: CI/CD pipelines, data pipelines, build
    workflows
  - infrastructure: Cloud resources, containers, K8s,
    and infrastructure-as-code
  - agent_skill: AI agent skills, prompts, system
    prompts, and LLM configurations
  - data: Datasets, data files, spreadsheets, reports
  - project: Entire projects, repositories, codebases
  - dependency: Packages, libraries, version management
  - security: Vulnerabilities, authentication, secrets
  - content: Text content, blog posts, marketing copy

**domain** -- choose ONE from:
  - web_frontend: Browser-based UI development (React,
    Vue, Angular, HTML/CSS)
  - backend_api: Server-side development, REST/GraphQL
    APIs, microservices
  - devops_infra: CI/CD, cloud infrastructure,
    containers, Kubernetes
  - data_ml: Data engineering, machine learning, AI
    model training, analytics
  - mobile: iOS, Android, cross-platform mobile apps
  - security: Application security, penetration testing,
    vulnerability management
  - database: Relational and NoSQL databases, query
    optimization
  - ai_agents: LLM applications, agent frameworks, RAG
    systems, prompt engineering
  - developer_tools: CLI tools, IDE extensions, code
    generation, developer productivity
  - testing_qa: Test automation, quality assurance
  - product_design: UI/UX design, product management,
    user research
  - systems: Operating systems, embedded systems,
    compilers
  - business_ops: Project management, marketing, sales,
    finance, legal

Respond ONLY with a JSON array. Each element:
{"id": "...", "primary_action": "...",
 "primary_object": "...", "domain": "..."}.
No explanations, no markdown fences.
\end{verbatim}
\end{quote}

\paragraph{User message format (per batch of 100 skills).}
\begin{quote}
\small
\begin{verbatim}
Tag these skills:
{id_1}|{name_1}: {description_first_200_chars}
{id_2}|{name_2}: {description_first_200_chars}
...
{id_100}|{name_100}: {description_first_200_chars}
\end{verbatim}
\end{quote}

\subsection{Consensus Clustering over Action--Object Combinations}
\label{app:clustering}

The action $\times$ object product space contains 182 possible combinations,
but the distribution is highly concentrated: the top 44 combinations account
for 80\% of all skills (14{,}324 of 17{,}810). We focus on these 44
combinations to discover stable groupings that seed the initial taxonomy
draft.

\paragraph{Embedding.}
Each combination is represented as the text \texttt{"\{action\} \{object\}"}
and encoded with Qwen3-Embedding-8B, yielding a 4{,}096-dimensional vector.

\paragraph{Multi-resolution clustering.}
We run $k$-means at five resolutions ($k \in \{5, 7, 10, 15, 20\}$)
with 20 random initializations each, and build a co-association matrix:
entry $(i, j)$ records the fraction of runs in which combinations $i$ and $j$
are assigned to the same cluster.

\paragraph{Strict consensus groups.}
Two combinations are linked if and only if they co-occur in \emph{all five}
resolutions (threshold = 5/5); that is, regardless of whether $k$ is 5 or
20, the pair is always assigned to the same cluster. Connected components
of this graph yield \textbf{10 stable groups} (25 combinations) and
\textbf{19 singletons} (Table~\ref{tab:stable-groups}).

The groups fall into two types: \emph{object-bound} groups, in which diverse
actions share a common object (e.g., G2: document\,$\times$\,doc,
generate\,$\times$\,doc, review\,$\times$\,doc); and \emph{action-bound}
groups, in which a single action spans multiple objects (e.g., G1:
implement\,$\times$\,code, implement\,$\times$\,api, implement\,$\times$\,data).
Object-bound groups outnumber action-bound groups 6 to 4; these
stable groups seed the initial taxonomy draft
(\S\ref{app:taxonomy-iteration}).

\begin{table}[t]
\centering
\caption{Consensus clustering: 10 stable groups (threshold = 5/5).
\emph{Binding} indicates whether members share a common object or action.
19 singletons (5{,}776 skills) are omitted for brevity.}
\label{tab:stable-groups}
\small
\begin{tabular}{clrc}
\toprule
\textbf{Group} & \textbf{Members (action $\times$ object)} & \textbf{Skills} & \textbf{Bind} \\
\midrule
G1  & implement\,$\times$\,code, impl.\,$\times$\,api, impl.\,$\times$\,data & 2{,}404 & action \\
G2  & document\,$\times$\,doc, generate\,$\times$\,doc, review\,$\times$\,doc & 1{,}309 & object \\
G3  & orchestrate\,$\times$\,agent, impl.\,$\times$\,agent, config.\,$\times$\,agent & 1{,}499 & object \\
G4  & configure\,$\times$\,infra, deploy\,$\times$\,infra & 637 & object \\
G5  & analyze\,$\times$\,code, analyze\,$\times$\,project & 627 & action \\
G6  & configure\,$\times$\,pipeline, impl.\,$\times$\,pipeline & 475 & object \\
G7  & orchestrate\,$\times$\,pipeline, deploy\,$\times$\,pipeline & 414 & object \\
G8  & design\,$\times$\,project, design\,$\times$\,api, design\,$\times$\,code & 441 & action \\
G9  & design\,$\times$\,agent, analyze\,$\times$\,agent, doc.\,$\times$\,agent & 444 & object \\
G10 & configure\,$\times$\,code, configure\,$\times$\,project & 298 & action \\
\bottomrule
\end{tabular}
\end{table}

\subsection{Iterative Taxonomy Design}
\label{app:taxonomy-iteration}

The final two-level taxonomy (Table~\ref{tab:taxonomy}) is the product
of an iterative, human-in-the-loop process.
The 10 stable groups identified by consensus clustering
(\S\ref{app:clustering}) were used to seed an initial draft taxonomy
with 7~Major categories and 21~Sub-categories.
Experts then iteratively reviewed stratified samples of 200~skills,
identifying structural ambiguities such as an over-broad
\emph{Documentation~\&~Knowledge} category, mixed classification axes
within Software Engineering, and scattered ML-related skills across
Data and SE\@.
Through successive rounds of review and revision, the taxonomy was
refined into the final \textbf{6~Major / 18~Sub} structure.

\begin{figure}[t]
\centering
\includegraphics[width=\linewidth]{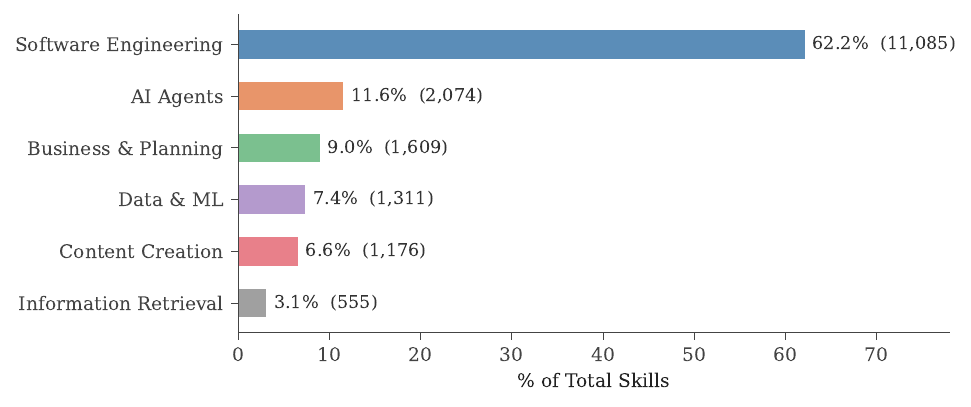}
\caption{Major-category distribution of the curated corpus of 17{,}810
skills prior to the data generation pipeline.
Software Engineering dominates (62.2\%), creating a
$20\times$ imbalance with the smallest category (Information
Retrieval, 3.1\%).}
\label{fig:category-dist}
\end{figure}

\begin{figure}[t]
\centering
\includegraphics[width=\linewidth]{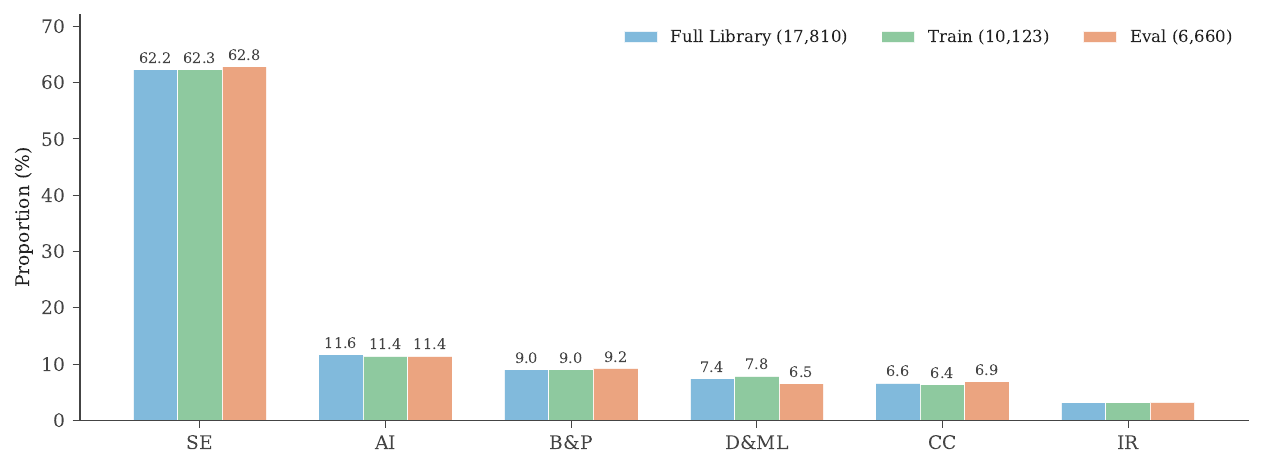}
\caption{Major-category distribution across data splits.
The three bars per category (full library, train, eval)
show near-identical proportions ($<$\,1\,pp deviation),
confirming that the disjoint split preserves the natural
category distribution.}
\label{fig:split-dist}
\end{figure}

\subsection{LLM-based Skill Assignment}
\label{app:llm-assignment}

While tag-based heuristic rules were effective for \emph{discovering}
the taxonomy structure, we found them insufficient for precise
\emph{assignment} of individual skills.
Three-axis tags (action, object, domain) capture only surface-level
attributes and cannot distinguish skills whose true purpose is
apparent only from the name and description.
For example, a skill tagged \texttt{implement / content / business\_ops}
is routed to Software Engineering by tag-based rules, although its
description reveals a marketing-campaign planner that belongs in
Business~\&~Planning.

To address these limitations, we classify all 17{,}810 skills in the
curated corpus using Claude Sonnet~4.6 via the Anthropic API\@.

\paragraph{System prompt.}
\begin{quote}
\small
\begin{verbatim}
You are a taxonomy classifier for AI agent skills.
Each skill is a reusable instruction file that extends
an LLM's capabilities. Given a skill's name and
description, assign it to exactly one (Major,
Sub-category) pair from the taxonomy below.

TAXONOMY:
## Software Engineering
   - Development / Analysis & Testing
   - Infrastructure & DevOps / Security
   - Version Control / Documentation
## AI Agents
   - Agent Development / Orchestration / Evaluation
## Data & ML
   - Data Engineering / Data Analysis / ML Development
## Content Creation
   - Writing & Text / Visual & Media
## Business & Planning
   - Business Analysis / Project Management
## Information Retrieval
   - Technical Search / General Search

CLASSIFICATION PRINCIPLE:
- Classify by the DOMAIN in which the skill's
  capability is used.
- Every skill extends an agent's capabilities,
  but classify by WHAT the extended capability
  is about, not the fact that an agent uses it.
- Technical docs (README, API docs) -> SE / Docs.
- Product planning (PRD, sprints, Jira)
  -> Business & Planning / Project Management.
- Pure business analysis (market research)
  -> Business & Planning / Business Analysis.
- Text/media as final product -> Content Creation.
- AI Agents is ONLY for the agent system itself
  (prompts, routing, MCP servers, evaluation).
- Information Retrieval is ONLY when the PRIMARY
  output is found/retrieved content.

OUTPUT: a JSON array, one object per skill.
  {"id": "...", "major": "...", "sub": "..."}
No markdown fences. No explanations.
\end{verbatim}
\end{quote}

\paragraph{User message format (per batch of 50 skills).}
\begin{quote}
\small
\begin{verbatim}
Classify these skills:

{id_1}|{name_1}: {description_first_300_chars}
{id_2}|{name_2}: {description_first_300_chars}
...
{id_50}|{name_50}: {description_first_300_chars}
\end{verbatim}
\end{quote}

\subsection{Human Validation of Assignment}
\label{app:human-validation}

To verify the quality of LLM-based assignment, a stratified random
sample of 200~skills is drawn from the classified corpus, preserving
the corpus-level distribution across all six Major categories.
Experts independently judge whether each skill's assigned
Major and Sub-category are appropriate.

The average accuracy across the reviewers is
\textbf{95.5\%} for major categories and \textbf{92.2\%} for
sub-categories. Full three-way agreement is reached on 91.0\%
(major) and 84.5\% (sub) of the 200 items.
Table~\ref{tab:human-validation} reports the per-category breakdown.

\begin{table}[h!]
\centering
\caption{Per-category accuracy of LLM-based taxonomy assignment,
averaged over independent reviewers on a stratified sample
of 200~skills.}
\label{tab:human-validation}
\small
\begin{tabular}{lrrr}
\toprule
\textbf{Major Category} & \textbf{$n$} & \textbf{Major Acc.} & \textbf{Sub Acc.} \\
\midrule
Software Engineering     & 120 & 97.5\% & 93.1\% \\
AI Agents                &  20 & 91.7\% & 85.0\% \\
Business \& Planning     &  20 & 96.7\% & 96.7\% \\
Data \& ML               &  15 & 82.2\% & 82.2\% \\
Content Creation         &  15 & 93.3\% & 93.3\% \\
Information Retrieval    &  10 & 100.0\% & 100.0\% \\
\midrule
\textbf{Total}           & \textbf{200} & \textbf{95.5\%} & \textbf{92.2\%} \\
\bottomrule
\end{tabular}
\end{table}

\section{Query Generation}
\label{app:query_generation}

\subsection{Query Generation Prompt}
  \label{app:query-gen-prompt}

  Each generation call receives the name and description of the sampled skill(s)
  as \texttt{\{skills\_text\}}, a random subset of 165 GAIA \cite{mialon2023gaia} validation questions as
  \texttt{\{seeds\_text\}}, and up to 30 previously generated queries for the
  same skill as \texttt{\{prev\_section\}} to suppress near-duplicate outputs.
  There is no system prompt. The entire instruction is issued as a single user turn.
  If the model judges the skill combination to be unrealistic, it outputs
  \texttt{None} and the combination is discarded.

  \paragraph{User prompt.}
  \begin{quote}
  \small
  \begin{verbatim}
  Write one realistic message that a user might send to an AI
  coding assistant.

  The message must naturally require ALL of the following
  skills to fulfill:

  {skills_text}

  Here are {N} examples of how real users talk to AI
  assistants. Notice the variety -- questions, commands,
  multi-step requests, short and long. Match this diversity
  of tone and structure:

  {seeds_text}

  ## Previously generated queries (DO NOT repeat or
  ## paraphrase these)
  {prev_queries}

  RULES:
  - Do NOT always start with "I'm" or "I need". Vary the
    opening: use questions ("How do I..."), commands
    ("Set up..."), descriptions ("Our team has..."), etc.
  - Do NOT mention skill names. The need must arise from
    the task description itself.
  - Do NOT explain, evaluate, or comment on the skills.
    Just write the user message.
  - The message must be standalone (no prior conversation
    context needed).
  - Your query must be DIFFERENT from any previously
    generated query listed above. Use a different scenario,
    domain, or framing.
  - If this skill combination makes no sense together in
    any realistic scenario, output exactly: None

  YOUR OUTPUT (one line only -- either a user message
  or None):
  \end{verbatim}
  \end{quote}

\subsection{Multi-Perspective LLM Review}
  \label{app:llm-review}

  Each query--skill pair that passes the leakage filter is evaluated by Claude
  Sonnet~4.6 using three independent reviewer prompts issued in separate API
  calls. Each prompt adopts a distinct evaluation persona—Skill Coherence,
  Query Quality, and Benchmark Discriminability—so that each dimension is
  assessed without anchoring bias from the others.
  A query is marked \textit{invalid} if two or more reviewers return an invalid
  verdict. A single invalid verdict routes the query to human expert review
  rather than discarding it outright.

  \paragraph{Reviewer 1 — Skill Coherence.}
  \begin{quote}
  \small
  \begin{verbatim}
  You are a benchmark quality reviewer evaluating skill-query alignment.

  SKILL(S):
  {skills_block}

  USER QUERY:
  {query}

  Does this query genuinely require the skill(s) listed above?
  - Is there a meaningful semantic connection between the skill
    description and the query?
  - If multiple skills are provided, does the query naturally
    require all of them?
  - Mark INVALID if the skill and query are unrelated or the
    combination is forced.

  Be conservative: only mark INVALID when clearly problematic.
  When in doubt, mark VALID.

  Respond in JSON only (no markdown):
  {"verdict": "valid" or "invalid", "reasoning": "1-2 sentences"}
  \end{verbatim}
  \end{quote}

  \paragraph{Reviewer 2 — Query Quality.}
  \begin{quote}
  \small
  \begin{verbatim}
  You are a benchmark quality reviewer evaluating query realism
  and specificity.

  SKILL(S):
  {skills_block}

  USER QUERY:
  {query}

  Is this a well-formed, realistic user query?
  - Is the request specific and answerable?
  - Could this plausibly come from a real user in a professional
    setting?
  - Is the content technically coherent?
  - Mark INVALID if the query is too vague, unrealistic, or
    technically incoherent.

  Be conservative: only mark INVALID when clearly problematic.
  When in doubt, mark VALID.

  Respond in JSON only (no markdown):
  {"verdict": "valid" or "invalid", "reasoning": "1-2 sentences"}
  \end{verbatim}
  \end{quote}

  \paragraph{Reviewer 3 — Benchmark Discriminability.}
  \begin{quote}
  \small
  \begin{verbatim}
  You are a benchmark quality reviewer evaluating whether a query
  can distinguish models that have access to the skill from those
  that do not.

  SKILL(S):
  {skills_block}

  USER QUERY:
  {query}

  Can this query discriminate between models with and without
  the skill?
  - Would a model lacking this specific skill fail to answer
    it well?
  - Is the query too generic -- answerable by any capable model
    without specialized skill knowledge?
  - Mark INVALID if the query can be answered adequately without
    the specific skill.

  Be conservative: only mark INVALID when clearly problematic.
  When in doubt, mark VALID.

  Respond in JSON only (no markdown):
  {"verdict": "valid" or "invalid", "reasoning": "1-2 sentences"}
  \end{verbatim}
  \end{quote}

\section{Functional-Equivalence and Graded-Relevance Audit}
\label{app:label-audit}

\paragraph{Binary-label correction.}
For each of the 6{,}660 skills in the pre-audit evaluation pool, which pairs with
4{,}997 queries, we retrieve the five nearest evaluation and training neighbors
with e5-mistral-7b-instruct. We retain candidate pairs with cosine similarity of
at least 0.80 and ask GPT-5.5, using full skill bodies, whether each pair is
functionally substitutable. Within the evaluation pool, union--find produces
equivalence groups, and we keep one representative per group while remapping
their qrels. We additionally remove evaluation skills with a train-pool
functional copy and drop queries that rely on them. Re-evaluating all 19
first-stage systems raises NDCG@10 by 3.29 points on average, with a standard
deviation of 0.33 and a range of 2.70--3.91.

\paragraph{Graded relevance.}
We also retain evaluation-internal functional equivalents as relevance level 1, assign the original seed skill level 2, and assign unrelated skills level 0. After removing train--evaluation leakage and any query that loses a seed, this variant contains 4{,}461 queries with 7{,}307 seed labels and 358 substitute labels. Table~\ref{tab:graded-relevance} compares binary and graded NDCG@10. Only 6.7\% of queries contain a substitute label. Across all 19 systems, graded minus binary NDCG@10 lies within 0.09 points (mean $-0.01$).

\begin{table}[h!]
\centering
\caption{Binary and graded NDCG@10 on the graded-relevance variant.}
\label{tab:graded-relevance}
\small
\begin{tabular}{lrrr}
\toprule
\textbf{Model} & \textbf{Binary} & \textbf{Graded} & \textbf{$\Delta$} \\
\midrule
SkillRet-Embedding-8B & 85.75 & 85.70 & $-0.05$ \\
SkillRouter-Embedding-0.6B & 72.91 & 72.97 & $+0.05$ \\
Qwen3-Embedding-8B & 62.86 & 62.77 & $-0.09$ \\
BM25 & 51.03 & 51.02 & $-0.00$ \\
e5-small-v2 & 44.07 & 44.09 & $+0.01$ \\
\bottomrule
\end{tabular}
\end{table}

\paragraph{Functional twins in training batches.}
The training set is multi-positive. Its 63,259 queries yield 127,190 positive
query-skill pairs, or approximately two skills per query. A functional twin
becomes an in-batch negative only when it appears in the same batch. Under
strict substitutability judgments and effective batch sizes of 384 and 80, this
occurs in approximately 0.3\% and 0.06\% of instances, and stays below 1.7\% and
0.4\% under a conservative bound that also counts partially overlapping pairs.
The corrected evaluation, cross-benchmark transfer, and Terminal-Bench, which
carries no relevance labels, provide three complementary checks that the gain
reflects retrieval quality rather than this benchmark's construction convention.


\section{Top-$k$ Reranking Depth Ablation}
\label{app:topk-ablation}

Table~\ref{tab:topk-ablation} reports NDCG@10 on the pre-audit split for three first-stage retrievers
across reranking depths $k \in \{10, 20, 50\}$ using Qwen3-Reranker-0.6B and
Qwen3-Reranker-8B.
Larger $k$ consistently improves NDCG@10 across all models and both rerankers.
We adopt $k{=}20$ in the main experiments as a practical trade-off between
performance and computational cost.

\begin{table*}[h!]
\centering
\caption{Pre-audit NDCG@10 at varying reranking depths $k \in \{10, 20, 50\}$
for two rerankers across three first-stage retrievers.
\textbf{Emb.\ Only} denotes the embedding-only baseline without reranking.}
\label{tab:topk-ablation}
\vspace{0.4em}
\small
\resizebox{\textwidth}{!}{%
\begin{tabular}{ll cccc}
\toprule
& & & \multicolumn{3}{c}{\textbf{NDCG@10 after Reranking}} \\
\cmidrule(lr){4-6}
\textbf{Reranker} & \textbf{First-Stage Model}
  & \textbf{Emb.\ Only} & \textbf{$k$=10} & \textbf{$k$=20} & \textbf{$k$=50} \\
\midrule
\multirow{3}{*}{Qwen3-Reranker-0.6B}
  & snowflake-arctic-embed-s & 53.00 & 58.50 & 61.90 & 65.40 \\
  & harrier-oss-v1-270m      & 61.17 & 65.40 & 67.54 & 69.90 \\
  & Qwen3-Embedding-8B       & 59.98 & 64.00 & 66.79 & 68.90 \\
\midrule
\multirow{3}{*}{Qwen3-Reranker-8B}
  & snowflake-arctic-embed-s & 53.00 & 58.00 & 61.30 & 64.70 \\
  & harrier-oss-v1-270m      & 61.17 & 65.10 & 67.04 & 69.70 \\
  & Qwen3-Embedding-8B       & 59.98 & 63.70 & 66.44 & 68.80 \\
\bottomrule
\end{tabular}%
}
\end{table*}


\section{Document Representation Ablation}
\label{app:doc-repr-ablation}

Table~\ref{tab:doc-repr-ablation} compares two document representation
strategies across three embedding models.
\textbf{Name+Desc} encodes only the skill name and description,
while \textbf{Full} encodes the complete document text including
the name, description, and Markdown body up to the model's maximum sequence length.
Full-text encoding consistently outperforms name-and-description only
across all models, with gains of 1.5--11.4 NDCG@10 points.

\begin{table}[h!]
\centering
\caption{Effect of document representation on pre-audit NDCG@10.}
\label{tab:doc-repr-ablation}
\vspace{0.4em}
\small
\begin{tabular}{l cc}
\toprule
\textbf{Model} & \textbf{Name+Desc} & \textbf{Full} \\
\midrule
snowflake-arctic-embed-s & 51.50 & 53.00 \\
harrier-oss-v1-270m      & 55.50 & 61.20 \\
Qwen3-Embedding-8B       & 48.60 & 60.00 \\
\bottomrule
\end{tabular}
\end{table}


\section{Model Maximum Sequence Lengths}
\label{app:max-seq-lengths}

Table~\ref{tab:max-seq-lengths} reports the maximum input sequence length for
each model evaluated in this work.
For each model, we use the maximum sequence length specified in the official model card
or documentation.
When no explicit limit is stated, we use the model's default context window.
Encoder-only embedding models are limited to 512 tokens, which truncates the majority of
skill documents in the corpus.
Decoder-only models support substantially longer contexts of 8K--32K tokens,
covering nearly all documents.
Detailed document length statistics are in Section~\ref{sec:skill-taxonomy-stats}.

\begin{table}[h!]
\centering
\caption{Maximum supported sequence length per embedding and reranking model.}
\label{tab:max-seq-lengths}
\vspace{0.4em}
\small
\begin{tabular}{llr}
\toprule
\textbf{Method} & \textbf{Model} & \textbf{Max Tokens} \\
\midrule
\multicolumn{3}{l}{\emph{Embedding}} \\
& bge-small-en-v1.5         & 512 \\
& e5-small-v2               & 512 \\
& snowflake-arctic-embed-s  & 512 \\
& bge-large-en-v1.5         & 512 \\
& e5-large-v2               & 512 \\
& F2LLM-v2-80M              & 40{,}960 \\
& harrier-oss-v1-270m       & 32{,}768 \\
& pplx-embed-v1-0.6b        & 32{,}768 \\
& Qwen3-Embedding-0.6B      & 32{,}768 \\
& jina-embeddings-v5-text-small & 8{,}192 \\
& harrier-oss-v1-0.6b       & 32{,}768 \\
& NV-Embed-v1               & 32{,}768 \\
& Octen-Embedding-8B        & 32{,}768 \\
& Qwen3-Embedding-8B        & 32{,}768 \\
& KaLM-Gemma3-12B           & 8{,}192 \\
\midrule
\multicolumn{3}{l}{\emph{Reranking}} \\
& jina-reranker-v2-base-multilingual & 1{,}024 \\
& Qwen3-Reranker-0.6B       & 32{,}768 \\
& Qwen3-Reranker-4B         & 32{,}768 \\
& Qwen3-Reranker-8B         & 32{,}768 \\
\bottomrule
\end{tabular}
\end{table}


\section{Retrieval Prompts for Each Evaluated Model}
\label{app:prompts-eval}

For each model, we follow the query/document prompts recommended in the
official model documentation, including model cards, READMEs, and reference implementations.
Three models deviate from their default prompts.

\begin{itemize}
  \item \textbf{Harrier-OSS and Qwen3-Embedding.}
    Most models use task-neutral prompts such as \texttt{query:} or \texttt{passage:},
    but both of these families default to a web search specific instruction.
    We replace it with a skill-retrieval instruction authored for this work,
    shown in Table~\ref{tab:eval-prompts}.
  \item \textbf{Octen-Embedding-8B.}
    Following the official README, we prepend \texttt{"- "} to
    each document to work around a known upstream tokenizer issue.\footnote{See
    \url{https://huggingface.co/Qwen/Qwen3-Embedding-8B/discussions/21}.}
  \item \textbf{Qwen3-Reranker.}
    The default web search instruction is replaced with a skill search instruction
    we authored, shown in Table~\ref{tab:eval-prompts}.
\end{itemize}

Table~\ref{tab:eval-prompts} lists the final query and document prompts used
for each model.

\begin{table*}[h!]
  \centering
  \caption{Query and document prompts for each model.
  \textit{none} denotes no prompt applied.
  $\dagger$~Prompt authored for this work.}
  \label{tab:eval-prompts}
  \vspace{0.4em}
  \renewcommand{\arraystretch}{1.3}
  \small
  \resizebox{\linewidth}{!}{%
  \begin{tabular}{p{5.5cm} p{9.0cm} p{1.8cm}}
  \toprule
  \textbf{Model} & \textbf{Query Prompt} & \textbf{Doc Prompt} \\
  \midrule
  \multicolumn{3}{l}{\emph{Embedding}} \\
  bge-small/large-en-v1.5
    & \texttt{Represent this sentence for searching relevant passages:} & \textit{none} \\
  snowflake-arctic-embed-s
    & \texttt{Represent this sentence for searching relevant passages:} & \textit{none} \\
  \addlinespace
  e5-small/large-v2
    & \texttt{query:} & \texttt{passage:} \\
  \addlinespace
  pplx-embed-v1-0.6b        & \texttt{Query:} & \texttt{Document:} \\
  jina-embeddings-v5-text-small & \texttt{Query:} & \texttt{Document:} \\
  \addlinespace
  harrier-oss-v1-270m/0.6b
    & \texttt{Instruct: Given a skill search query, retrieve relevant skills that match the query\textbackslash{}nQuery:}$^\dagger$ & \textit{none} \\
  Qwen3-Embedding-0.6B/8B
    & \texttt{Instruct: Given a skill search query, retrieve relevant skills that match the query\textbackslash{}nQuery:}$^\dagger$ & \textit{none} \\
  \addlinespace
  F2LLM-v2-80M
    & \texttt{Instruct: Given a question, retrieve passages that can help answer the question.\textbackslash{}nQuery:} & \textit{none} \\
  KaLM-Gemma3-12B
    & \texttt{Instruct: Given a query, retrieve documents that answer the query\textbackslash{}nQuery:} & \textit{none} \\
  \addlinespace
  NV-Embed-v1               & \textit{none} & \textit{none} \\
  Octen-Embedding-8B        & \textit{none} & \texttt{- } \\
  \midrule
  \multicolumn{3}{l}{\emph{Reranking}} \\
  jina-reranker-v2-base-multilingual & \textit{none} & \textit{none} \\
  Qwen3-Reranker-0.6B/4B/8B & \texttt{Given a skill search query, judge whether the skill document is relevant and useful for the query}$^\dagger$ & \textit{none} \\
  \bottomrule
  \end{tabular}%
  }
  \end{table*}


\section{Fine-tuning Base Model Selection}
\label{app:base-model-selection}

To select the base model for fine-tuning, we compared fine-tuning
Qwen3-Embedding-0.6B against fine-tuning harrier-oss-v1-0.6b,
which is itself a derivative of Qwen3-Embedding.
Table~\ref{tab:base-model-selection} shows that both fine-tuned
variants achieve nearly identical performance across all metrics,
with differences well within noise.
We therefore choose Qwen3-Embedding as the fine-tuning base to avoid
double fine-tuning and to maintain a cleaner experimental provenance.
The same rationale applies at the 8B scale, where we fine-tune
Qwen3-Embedding-8B in preference to Octen-Embedding-8B,
which is also Qwen3-Embedding-based.

\begin{table}[h!]
\centering
\caption{Fine-tuning base model comparison at 0.6B scale on the pre-audit split.
(ft) denotes fine-tuned on SkillRet training data.}
\label{tab:base-model-selection}
\vspace{0.4em}
\small
\resizebox{\columnwidth}{!}{%
\begin{tabular}{l ccc ccc ccc}
\toprule
& \multicolumn{3}{c}{\textbf{NDCG}} & \multicolumn{3}{c}{\textbf{Recall}} & \multicolumn{3}{c}{\textbf{Completeness}} \\
\cmidrule(lr){2-4} \cmidrule(lr){5-7} \cmidrule(lr){8-10}
\textbf{Model}
& \textbf{@5} & \textbf{@10} & \textbf{@15}
& \textbf{@5} & \textbf{@10} & \textbf{@15}
& \textbf{@5} & \textbf{@10} & \textbf{@15} \\
\midrule
harrier-oss-v1-0.6b      & 64.24 & 66.55 & 67.54 & 67.06 & 73.09 & 76.27 & 50.57 & 57.37 & 61.12 \\
harrier-oss-v1-0.6b (ft) & 75.80 & 78.34 & 79.34 & 79.65 & 86.13 & 89.28 & 66.72 & 76.03 & 80.83 \\
\midrule
Qwen3-Embedding-0.6B     & 56.23 & 58.35 & 59.34 & 59.29 & 64.89 & 68.06 & 41.98 & 47.27 & 50.31 \\
SkillRet-Embedding-0.6B (ft) & 75.57 & 78.03 & 78.87 & 79.15 & 85.42 & 88.09 & 65.96 & 75.09 & 79.03 \\
\bottomrule
\end{tabular}%
}
\end{table}


\section{SkillRet Fine-tuning Details}
\label{app:finetuning-details}

We fine-tune all SkillRet models on the released training split, comprising
10{,}123 skills and 63{,}259 synthetic queries yielding 127{,}190 positive
query--skill pairs. Training and evaluation skills are disjoint.

\paragraph{Embedding models.}
We fine-tune Qwen3-Embedding-0.6B and Qwen3-Embedding-8B using MultipleNegativesRankingLoss.
Each query is paired with one positive skill document per training instance,
so a query with multiple ground-truth skills contributes multiple pairs,
with remaining in-batch examples serving as negatives. Skill documents are
encoded as \texttt{name | description | skill\_md}, matching the evaluation
document representation. We apply the same skill-retrieval query instruction
used in evaluation (Table~\ref{tab:eval-prompts}) to anchor queries during
training. Both models are trained for one epoch with maximum sequence length
8192, learning rate $2\times10^{-5}$, warmup ratio 0.1, bf16 precision, and
gradient checkpointing on 4 GPUs. The 0.6B model uses per-device batch size 96, 
effective batch 384, while the 8B model uses per-device batch size 20, effective batch 80.

\paragraph{Reranker model.}
We fine-tune Qwen3-Reranker-0.6B using the same yes/no token scoring interface
used at inference time. For each query--document pair, the model receives a
chat-formatted prompt containing the skill-search instruction, query, and
candidate skill document, and is trained with binary cross-entropy on the
probability of the ``yes'' token versus the ``no'' token. Positive pairs come from the ground-truth
query--skill labels. For negatives, we mine hard negatives using the fine-tuned
SkillRet-Embedding-0.6B retriever. For each query, we retrieve the top 60 candidates,
skip the top 20 near-neighbor candidates, and use up to 7 remaining non-relevant
candidates, filling any missing slots with random negatives. The reranker is trained for one epoch with maximum sequence length 8192,
learning rate $2\times10^{-5}$, warmup ratio 0.1, bf16 precision, and
gradient checkpointing on 8 GPUs. Per-device batch size is 96, effective batch 768.

\section{MTEB Retrieval vs.\ \textsc{SkillRet} Performance}
\label{app:mteb-comparison}

Table~\ref{tab:mteb-vs-skillret} lists MTEB Retrieval scores alongside \textsc{SkillRet} NDCG@10
for all evaluated models, sorted by MTEB Retrieval score in descending order.
A moderate positive correlation is visible in the overall trend,
yet notable exceptions appear in both directions.
KaLM-Gemma3-12B, for instance, leads on MTEB at 75.66
but achieves only 58.95 on corrected \textsc{SkillRet}, the largest drop among all models.
Conversely, some models with low MTEB scores remain highly competitive on \textsc{SkillRet}.
harrier-oss-v1-0.6b ranks 4th on MTEB at 70.75 yet achieves the best off-the-shelf score
on \textsc{SkillRet} at 70.26, and encoder-only models with as few as 33M parameters
reach corrected \textsc{SkillRet} scores in the mid-50s, comparable to
NV-Embed-v1 at 7B which scores 57.03 despite a substantially higher MTEB score of 53.98.
Together, these patterns suggest that skill retrieval is a distinct task from general information retrieval,
requiring models to identify specific capability signals within long, multi-sentence queries.

\begin{table}[h!]
\centering
\caption{MTEB Retrieval score vs.\ \textsc{SkillRet} NDCG@10.
Models sorted by MTEB Retrieval score in descending order.}
\label{tab:mteb-vs-skillret}
\vspace{0.4em}
\small
\begin{tabular}{lc cc}
\toprule
\textbf{Model} & \textbf{Params}
  & \textbf{MTEB Retr.} & \textbf{SkillRet NDCG@10} \\
\midrule
KaLM-Gemma3-12B              & 12B  & 75.66 & 58.95 \\
Octen-Embedding-8B           & 8B   & 71.61 & 65.93 \\
Qwen3-Embedding-8B           & 8B   & 70.88 & 63.64 \\
harrier-oss-v1-0.6b          & 0.6B & 70.75 & 70.26 \\
harrier-oss-v1-270m          & 270M & 66.38 & 64.56 \\
jina-embeddings-v5-text-small & 0.6B & 64.88 & 62.96 \\
Qwen3-Embedding-0.6B         & 0.6B & 64.65 & 61.94 \\
NV-Embed-v1                  & 7B   & 53.98 & 57.03 \\
F2LLM-v2-80M                 & 80M  & 50.13 & 48.22 \\
e5-large-v2                  & 335M & 40.06 & 53.41 \\
snowflake-arctic-embed-s     & 33M  & 39.84 & 56.39 \\
e5-small-v2                  & 118M & 39.38 & 44.66 \\
bge-large-en-v1.5            & 335M & 39.00 & 59.00 \\
bge-small-en-v1.5            & 33M  & 36.26 & 54.51 \\
\bottomrule
\end{tabular}
\end{table}

\section{Per-Sub-category Retrieval Performance}
\label{app:per-sub-category}

Table~\ref{tab:per-sub-category} provides a complete breakdown of
NDCG@10 and Recall@10 across all 18 Sub-categories for both base
and fine-tuned Qwen3-Embedding models.
Sub-categories are grouped by Major category and sorted by fine-tuned
8B NDCG@10 within each group.
This table supports the intra-Major hard-negative analysis in
\S\ref{sec:per-category}.

\begin{table*}[h!]
\centering
\caption{Per-Sub-category NDCG@10 and Recall@10 on the corrected evaluation set for base and fine-tuned
Qwen3-Embedding models. Sub-categories grouped by Major category.
A query is counted in every Sub-category represented among its gold skills, and
$n$ is the resulting number of evaluation queries per Sub-category.}
\label{tab:per-sub-category}
\vspace{0.3em}
\footnotesize
\setlength{\tabcolsep}{3pt}
\begin{tabular}{@{}ll r cc cc cc cc@{}}
\toprule
& & & \multicolumn{4}{c}{\textbf{NDCG@10}} & \multicolumn{4}{c}{\textbf{Recall@10}} \\
\cmidrule(lr){4-7} \cmidrule(lr){8-11}
\textbf{Major} & \textbf{Sub-category} & \textbf{$n$}
  & \textbf{0.6B} & \textbf{0.6B-ft} & \textbf{8B} & \textbf{8B-ft}
  & \textbf{0.6B} & \textbf{0.6B-ft} & \textbf{8B} & \textbf{8B-ft} \\
\midrule
\multirow{6}{*}{\rotatebox{90}{Software Eng.}}
  & Infra.\ \& DevOps     &  818 & 64.4 & 83.4 & 66.2 & 89.3 & 68.1 & 90.0 & 71.7 & 95.0 \\
  & Documentation         &  367 & 56.8 & 80.5 & 55.8 & 87.0 & 60.5 & 87.2 & 60.7 & 92.7 \\
  & Development           & 1509 & 61.5 & 81.1 & 65.6 & 86.9 & 66.3 & 86.9 & 71.4 & 92.6 \\
  & Security              &  268 & 63.3 & 81.3 & 68.2 & 86.2 & 67.2 & 87.8 & 73.2 & 91.8 \\
  & Analysis \& Testing   &  859 & 54.9 & 77.9 & 59.7 & 84.9 & 59.1 & 84.6 & 64.4 & 91.9 \\
  & Version Control       &  316 & 48.3 & 75.2 & 50.2 & 82.8 & 52.8 & 81.6 & 55.6 & 90.8 \\
\midrule
\multirow{3}{*}{\rotatebox{90}{AI Ag.}}
  & Agent Evaluation      &   95 & 52.3 & 76.3 & 52.2 & 84.6 & 56.1 & 80.9 & 57.5 & 89.1 \\
  & Agent Development     &  422 & 48.0 & 74.4 & 49.5 & 83.5 & 52.5 & 81.0 & 54.9 & 91.2 \\
  & Agent Orchestration   &  275 & 41.9 & 68.5 & 41.5 & 76.0 & 47.2 & 72.1 & 45.6 & 81.8 \\
\midrule
\multirow{2}{*}{\rotatebox{90}{B\&P}}
  & Business Analysis     &  334 & 57.0 & 81.4 & 59.5 & 86.9 & 60.8 & 87.1 & 64.0 & 92.0 \\
  & Project Mgmt.         &  329 & 51.8 & 78.2 & 50.8 & 83.3 & 58.4 & 85.6 & 56.9 & 91.2 \\
\midrule
\multirow{3}{*}{\rotatebox{90}{D\&ML}}
  & Data Analysis         &  134 & 67.8 & 90.1 & 71.7 & 92.7 & 68.8 & 93.7 & 73.3 & 95.6 \\
  & ML Development        &  163 & 70.5 & 86.7 & 75.8 & 89.6 & 72.3 & 90.3 & 78.1 & 93.5 \\
  & Data Engineering      &  180 & 65.6 & 81.7 & 69.0 & 89.2 & 71.2 & 87.1 & 73.5 & 94.5 \\
\midrule
\multirow{2}{*}{\rotatebox{90}{CC}}
  & Writing \& Text       &  299 & 53.8 & 84.1 & 55.0 & 89.5 & 57.2 & 89.4 & 59.8 & 95.6 \\
  & Visual \& Media       &  214 & 58.5 & 81.0 & 59.4 & 85.6 & 64.4 & 87.9 & 64.6 & 93.1 \\
\midrule
\multirow{2}{*}{\rotatebox{90}{IR}}
  & General Search        &  154 & 48.2 & 76.9 & 48.2 & 83.5 & 50.2 & 81.8 & 52.6 & 90.6 \\
  & Technical Search      &   98 & 37.2 & 66.4 & 34.0 & 72.0 & 39.6 & 72.1 & 35.7 & 81.6 \\
\bottomrule
\end{tabular}
\end{table*}

\section{Qualitative Visualization of Sentence Erasure Importance}
\label{app:loo-viz}

\begin{figure}[t]
    \centering
    \includegraphics[width=0.6\linewidth]{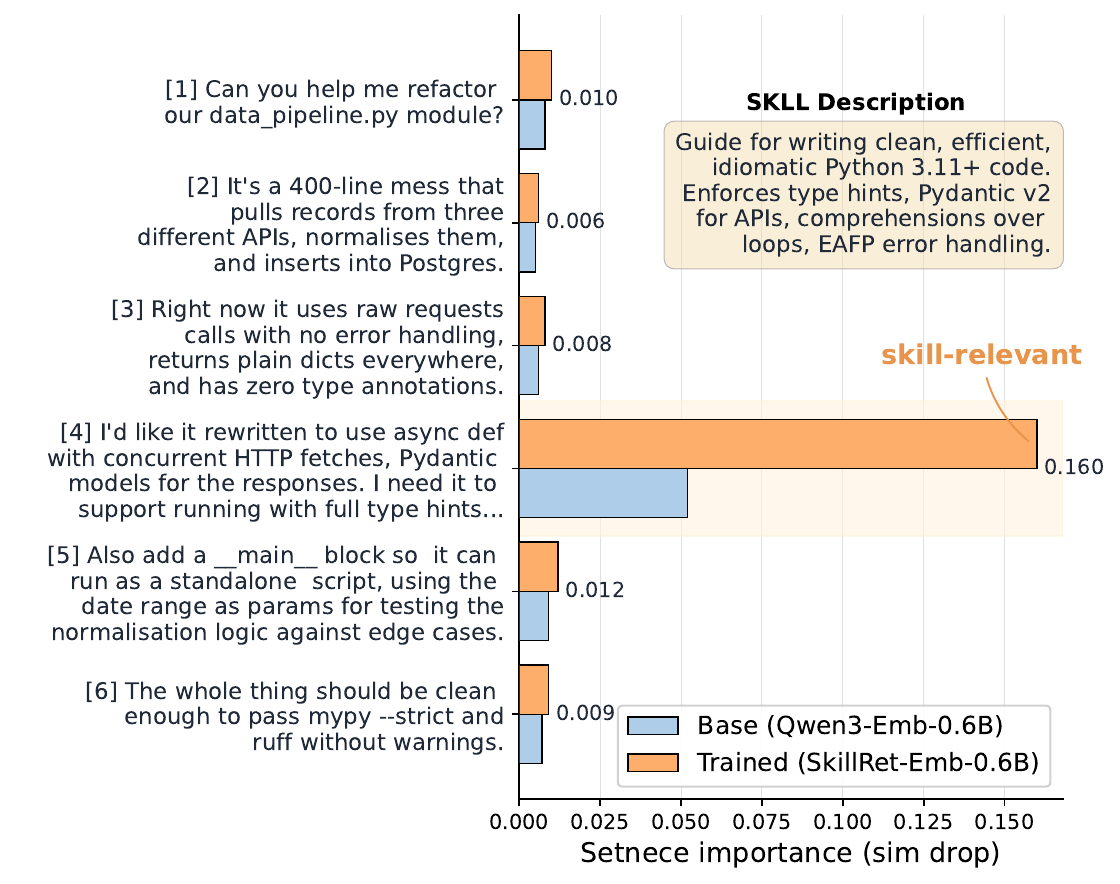}
    \caption{%
      Sentence-level erasure importance for an example query.
      Each bar shows the similarity drop after replacing a sentence with
      \texttt{[MASK]}.
      The trained model concentrates more importance on the skill-relevant sentence,
      whereas the base model assigns importance more diffusely across the query.
    }
    \label{fig:loo-ex}
\end{figure}

To complement the aggregate masking results in Table~\ref{tab:query-mask-ablation-main},
we provide a qualitative example of the sentence-level erasure analysis in
Fig.~\ref{fig:loo-ex}.
For each sentence in the query, we measure the similarity drop after replacing that
sentence with \texttt{[MASK]}.
A larger drop indicates that the sentence contributes more strongly to retrieving the
gold skill.

\DIFaddbegin

\section{Multi-Skill Retrieval by Category}
\label{app:multi-skill}

Table~\ref{tab:multi-skill-category} reports NDCG@10 for Qwen3-Embedding-0.6B
and SkillRet-Embedding-0.6B on the corrected set. In five of the six major
categories the fine-tuning gain widens monotonically from single-skill to
three-skill queries, and the categories whose base score falls furthest are the
ones whose gain grows most. Data \& ML falls 41.6 points from $k{=}1$ to $k{=}3$
and its gain widens by 29.2.

Information Retrieval is the exception because the task is already hard at
$k{=}1$. Its base score of 44.5, 43.5, and 44.6 barely moves, so the gain stays
near its $k{=}1$ value of 32.3 points instead of growing. Its skills are the
shortest in the corpus and it is the smallest major category, so difficulty
comes from how interchangeable the documents are rather than from how many are
required.

\begin{table*}[h!]
\centering
\caption{NDCG@10 by major category and number of required skills, shown as base $\rightarrow$ fine-tuned 0.6B. A query is counted in every major category among its gold skills, and NDCG is computed over its full gold set.}
\label{tab:multi-skill-category}
\small
\begin{tabular}{lccc}
\toprule
\textbf{Category} & \textbf{$k{=}1$} & \textbf{$k{=}2$} & \textbf{$k{=}3$} \\
\midrule
AI Agents & 48.8 $\rightarrow$ 71.2 & 48.6 $\rightarrow$ 74.5 & 39.2 $\rightarrow$ 72.3 \\
Business \& Planning & 68.5 $\rightarrow$ 83.7 & 49.9 $\rightarrow$ 79.9 & 40.2 $\rightarrow$ 72.2 \\
Content Creation & 73.7 $\rightarrow$ 88.4 & 52.6 $\rightarrow$ 82.8 & 40.4 $\rightarrow$ 75.7 \\
Data \& ML & 87.8 $\rightarrow$ 90.4 & 63.7 $\rightarrow$ 86.0 & 46.2 $\rightarrow$ 78.0 \\
Information Retrieval & 44.5 $\rightarrow$ 76.8 & 43.5 $\rightarrow$ 70.2 & 44.6 $\rightarrow$ 72.0 \\
Software Engineering & 79.5 $\rightarrow$ 85.9 & 54.1 $\rightarrow$ 79.8 & 40.1 $\rightarrow$ 73.4 \\
\bottomrule
\end{tabular}
\end{table*}

\section{Retrieval Latency and Throughput}
\label{app:efficiency}

We time each architecture on 100 fixed queries after 10 warm-up iterations on a
single NVIDIA B200, using HuggingFace Transformers, FlashAttention-2, and
bfloat16. Embedding latency uses batch size 1. Throughput is measured over 1,024
queries at the largest batch size each model fits on the same GPU, determined by
stress test and listed in Table~\ref{tab:embedding-latency}, so throughput is
comparable within a model rather than across models of different size.Reranking latency, shown in Table~\ref{tab:reranker-latency}, covers one
    query with 20 candidate documents. Models whose batch size
reaches 20 score all candidates in one pass, while Qwen3-Reranker-8B fits only
four at a time and therefore needs five passes, which the reported latency
includes. BM25 runs on CPU and is shown for reference. Fine-tuned and
SkillRouter checkpoints share the same runtime as their corresponding Qwen
architectures, since fine-tuning does not change the architecture.

\begin{table}[h!]
\centering
\caption{Embedding latency at batch size 1 and batched throughput.}
\label{tab:embedding-latency}
\small
\begin{tabular}{lcrrr}
\toprule
\textbf{Model} & \textbf{Params} & \textbf{Batch} & \textbf{Latency (ms)} & \textbf{Queries/s} \\
\midrule
BM25 (CPU)            & --   & --   & 0.73  & 1,405 \\
bge-small-en-v1.5     & 33M  & 1024 & 6.57  & 624 \\
bge-large-en-v1.5     & 335M & 1024 & 13.57 & 142 \\
Qwen3-Embedding-0.6B  & 0.6B & 16   & 21.25 & 381 \\
Qwen3-Embedding-8B    & 8B   & 8    & 26.25 & 129 \\
KaLM-Gemma3-12B       & 12B  & 4    & 60.58 & 43 \\
\bottomrule
\end{tabular}
\end{table}

\begin{table}[h!]
\centering
\caption{Mean latency to rerank one query's top-20 documents.}
\label{tab:reranker-latency}
\small
\begin{tabular}{lcrr}
\toprule
\textbf{Reranker} & \textbf{Params} & \textbf{Batch} & \textbf{Latency (ms)} \\
\midrule
jina-reranker-v2     & 0.3B & 1024 & 132.1 \\
Qwen3-Reranker-0.6B  & 0.6B & 32   & 831.4 \\
Qwen3-Reranker-8B    & 8B   & 4    & 1,875.5 \\
\bottomrule
\end{tabular}
\end{table} 

\section{Broader Impacts}
\label{app:broader-impacts}

\textsc{SkillRet} is intended to support research on reliable skill retrieval for LLM agents. By isolating retrieval quality from downstream execution, it provides a controlled benchmark for studying how well models select relevant procedural knowledge from large skill libraries. This may help reduce context cost, improve reproducibility, and diagnose retrieval failures across domains.
At the same time, strong retrieval performance does not imply safe or correct end-to-end agent behavior. Retrieved skills may be outdated, unsafe, misapplied, or incorrectly composed with other skills. Therefore, \textsc{SkillRet} should not be used as evidence that a deployed agent system is safe or reliable.
The dataset is derived from public GitHub-hosted skills and synthetic queries. It is intended for retrieval evaluation and model development, not for profiling individual authors, inferring personal attributes, or certifying downstream agent safety. Practical deployments should include additional safeguards such as provenance checks, permission controls, sandboxing, and human oversight for high-impact actions.

\end{document}